%% file: 00-main.tex
\documentclass[sigconf, nonacm]{acmart}

\usepackage{amsmath}
\usepackage{float}
\usepackage{graphicx}
\usepackage{textcomp}
\usepackage{xcolor}
\usepackage{algorithmic}
\usepackage[linesnumbered,ruled,vlined]{algorithm2e}
\usepackage{comment}
\usepackage{blindtext}
\usepackage{siunitx}
\usepackage{subcaption}
\usepackage{balance}
\usepackage{bm}
\usepackage{multirow}
\usepackage{booktabs}
\usepackage{url}
\usepackage{threeparttable}
\usepackage[T1]{fontenc}
\usepackage{fix-cm}
\usepackage{makecell}
\usepackage{tabularx}

\newcommand{\solution}{\textit{STELLA}}
\begin{document}

\title{STELLA: Efficient Sensor-to-LLM Translation for On-Device Human Activity Recognition}


\author{Nirhoshan Sivaroopan}
\affiliation{%
  \institution{The University of Sydney}
  \country{Australia}
  }

\author{Albert Zomaya}
\affiliation{%
  \institution{The University of Sydney}
  \country{Australia}
  }

\author{Kanchana Thilakarathna}
\affiliation{%
  \institution{The University of Sydney}
  \country{Australia}
  }








\begin{abstract}

\input{Files/Abstract}
\end{abstract}





\maketitle


\input{Files/Introduction-v3}

\input{Files/Related_work}
\input{Files/Methodology-v2}
\input{Files/Results-v4}

\input{Files/Discussion}
\input{Files/Conclusion}

\begin{acks}
This research was partially funded by the Australian Research Council Industrial Transformation Research Hub for Future Digital Manufacturing (IH230100013).
\end{acks}

\bibliographystyle{ACM-Reference-Format}
\bibliography{99-base}

\appendix

\input{Files/Appendix}

\end{document}

%% file: Files/Abstract.tex
HAR is increasingly expected to run continuously on edge devices, yet recent LLM-based methods remain hard to deploy: raw sensor prompts are long, cloud inference adds latency and privacy risk, and fine-tuned LLM pipelines turn general-purpose models into task-specific classifiers. We present \solution{}, an efficient sensor-to-LLM translation framework for on-device HAR that shifts the burden from LLM adaptation to sensor tokenization. A lightweight hierarchical tokenizer compresses an entire multi-channel inertial window into a fixed set of compact latent sensor tokens, which are projected into the embedding space of a frozen pretrained LLM and combined with a natural-language prompt for label scoring. This preserves activity-relevant temporal and cross-channel structure while keeping LLM-side computation predictable across sensor configurations. \solution{} also supports on-device personalization, adapting only the lightweight tokenizer on small amounts of user-specific labelled data and augmenting inference with a local retrieval context, keeping the LLM, user data, and retrieval on device. Across seven public HAR datasets and eight benchmark settings, \solution{} achieves new state-of-the-art performance, improving over prior methods by up to 11.83\% F1; on-device personalization yields up to a further 21.91\% F1 as user data accumulates after deployment. \solution{} also outperforms representative time-series tokenizers under the same LLM pipeline and achieves real-time inference under practical mobile and edge budgets, showing that efficient sensor tokenization is a practical path toward accurate, private, and personalized LLM-based HAR on edge devices.

%% file: Files/Introduction-v3.tex
\section{Introduction}

Human Activity Recognition (HAR) from IMU sensors is a core building block for mobile and ubiquitous computing applications, including health monitoring~\cite{liaqat2019wearbreathing}, rehabilitation~\cite{panwar2019rehab}, fitness tracking~\cite{bachlin2009wearable}, smart environments, and industrial sensing~\cite{grzeszick2017deep}. Modern smartphones, smartwatches, and wearable devices continuously capture accelerometer, gyroscope, and magnetometer signals, enabling fine-grained understanding of human behaviour without explicit user input. As these devices become increasingly capable, HAR systems are expected to be not only accurate, but also efficient enough for real-time execution on resource-constrained edge devices. They must also adapt after deployment: the same activity can produce different motion signatures across users due to differences in gait, speed, posture, device placement, and execution style~\cite{sivaroopan2025rag}.

For many years, deep learning has been the dominant paradigm for sensor-based HAR. CNN~\cite{zeng2014convolutional}, recurrent~\cite{guan2017ensembles}, and attention-based models~\cite{ahmad2023alae} have produced strong results by learning discriminative patterns directly from labelled sensor windows. More recently, however, Large Language Models (LLMs) have reshaped a wide range of machine learning problems by providing strong sequence modelling ability, semantic priors, and flexible reasoning over task descriptions and label spaces~\cite{sivaroopan2025rag}. These capabilities are especially attractive for HAR: activity labels such as \textit{running}, \textit{cycling}, or \textit{ironing} are not arbitrary class IDs, but semantically meaningful descriptions of human motion. This opens a new opportunity: instead of treating HAR purely as closed-set signal classification, we can formulate activity recognition as a sensor-to-language reasoning problem, where sensor observations are interpreted in the context of natural-language activity descriptions~\cite{ji2024hargpt}.

Early LLM-based HAR studies~\cite{ji2024hargpt, sivaroopan2025rag, hota2025evaluating} confirm the promise of this direction: LLMs can label activities from sensor-derived prompts, draw on pretrained knowledge of human behaviour, and generalize across several HAR settings. Their successes, however, expose a deeper obstacle that becomes decisive on-device, and it stems from a single mismatch: LLMs are built to reason over short sequences of discrete language tokens, whereas an activity window for time-series sensor data stream is a long, multi-channel stream of continuous floating-point values. This mismatch surfaces as three concrete problems. \emph{(i) Serializing a window as text is both ineffective and expensive.} Numerical arrays are foreign to text tokenizers, which fragment each value into many sub-tokens that the model reasons over poorly~\cite{spathis2024first}; and a single window already holds hundreds to thousands of scalars across channels, so the resulting prompt is very long. Such prompts inflate the LLM's quadratic attention cost and KV-cache memory, driving latency and footprint past what edge compute and memory budgets allow. \emph{(ii) Adapting the LLM itself trades generality for a task-specific model.} Fine-tuning or surgically rewriting the LLM so it ingests sensors directly turns a general-purpose model into a HAR-specific classifier; on the edge this is self-defeating, because each new sensing task would then need its own specialized model, and hosting several such models exceeds the memory and compute of phones and wearables, which can realistically run only one local LLM. \emph{(iii) Offloading to the cloud breaks the deployment guarantees.} Sending sensor data to a hosted LLM avoids both on-device costs but reintroduces network dependency and unbounded latency that undermine real-time operation~\cite{sivaroopan2025rag}, while streaming behaviourally sensitive sensor data off-device raises clear privacy concerns~\cite{yang2024privacy}.

Together, these obstacles indicate where on-device LLM-based HAR should go. Edge platforms increasingly host a single small, general-purpose LLM as a shared system-level service for text, voice, and multimodal reasoning, and the natural way to add HAR is to admit sensors as one more modality for this shared model, rather than compressing it, rewriting it, duplicating it per task, or shipping data to the cloud. What is missing is an efficient \emph{sensor-to-LLM interface}: a compact front-end that translates a raw multi-channel window into a short, language-model-compatible representation the frozen LLM can consume, preserving activity-relevant temporal and cross-channel structure while keeping the LLM general and reusable. In other words, efficient on-device LLM-based HAR should not begin by compressing or rewriting the LLM; it should begin by learning to tokenize sensor data well~\cite{li2025sensorllm}.

In this paper, we propose \solution{}:\textbf{S}ensor \textbf{T}okenization for \textbf{E}dge \textbf{LL}M \textbf{A}ctivity Recognition, an efficient sensor-to-LLM translation framework for on-device HAR. The central idea is a lightweight sensor tokenizer that converts a raw multi-channel window into a compact set of latent sensor tokens; a shallow projection then maps these tokens into the embedding space of a pretrained LLM, where they are combined with a natural-language task prompt so the LLM can score candidate activity labels by language-model likelihood. Rather than processing raw numerical streams, the LLM thus receives a short, semantically rich token sequence, while the learned front-end performs all sensor-specific work. This design targets three goals: inference cost should be governed by the learned token budget rather than the raw window length; the tokens should preserve multi-scale motion structure, capturing both local primitives such as step transients and longer-range dynamics such as periodic gait; and the LLM should remain a frozen, reusable reasoning backend rather than being modified for a single HAR configuration. The same design also supports on-device personalization: after deployment, only the lightweight sensor-side module is adapted using a small amount of user-specific labelled data, while the LLM remains frozen and reusable.

We evaluate \solution{} on seven public HAR datasets across eight benchmark settings, spanning 6 to 60 sensor channels and 6 to 18 activity classes and covering locomotion, daily-living activities, exercise, and industrial gestures. \solution{} sets new state-of-the-art (SOTA) performance on every benchmark, improving over the strongest prior methods by approximately 5.5 accuracy points and 3.6 F1 points on average, with the largest gains on low-channel single-IMU datasets and consistent improvements even on near-saturated benchmarks where prior methods already perform strongly. With on-device personalization, adapting only the lightweight tokenizer together with a local retrieval context lifts the hardest subject from 71\% to 93\% F1 while adding under 1\,ms of retrieval overhead.

Two further studies establish why \solution{} works and is deployable. First, under a fixed frozen-LLM pipeline and a matched token budget, the \solution{} tokenizer outperforms representative SOTA time-series encoders, showing that the gains come from a learned sensor-to-LLM translation layer rather than from generic time-series encoding. Second, an on-device feasibility analysis shows that \solution{} achieves real-time inference within practical mobile and edge budgets: because each window is compressed to a fixed token set, LLM-side computation stays predictable across channel counts and window lengths, and because inference and personalization run locally, sensor data and user-specific context never leave the device. 
In summary, this paper makes the following contributions:
\vspace{-2mm}
\noindent
\begin{itemize}
    \item \textbf{Reframing efficient on-device LLM-based HAR as a sensor-tokenization problem.} Prior work either prompts a cloud LLM or compresses/fine-tunes the LLM into a task-specific model, trading away privacy, predictable latency, or general-purpose reusability. We instead argue that the real bottleneck is the \emph{sensor-to-LLM interface}, and solve the efficiency problem entirely on the sensor side, so HAR can be attached to a shared local LLM as one more input modality rather than as a separate, fine-tuned model stack.

    \item \textbf{A hierarchical tokenizer that decouples LLM cost from sensor configuration.} The \solution{}-Tower compresses an \emph{entire} multi-channel window \emph{jointly} into a fixed set of $N{=}16$ latent tokens, via local temporal mixing, progressive downsampling, and multi-scale aggregation, rather than tokenizing each channel separately. Because the token budget is fixed, LLM-side compute and latency become invariant to channel count and window length ( up to $262.5\times$ fewer tokens than per-scalar encoding).

    \item \textbf{Lightweight, private on-device personalization.} After deployment, \solution{} adapts only the tokenizer ($\sim$10\% of parameters) to a user and augments inference with a local retrieval memory that is re-encoded as the tokenizer evolves. This improves the hardest subjects by up to $+21.91$ F1 with $<1$\,ms retrieval overhead, while the LLM stays frozen and all user data and context remain on-device.

    \item \textbf{State-of-the-art accuracy that remains deployable on-device.} Across seven datasets and eight benchmark settings, \solution{} achieves new SOTA (up to $+11.83$ F1) using only a frozen GPT-2 backbone, surpassing methods built on far larger LLMs. Under a matched pipeline and token budget, it likewise outperforms representative time-series tokenizers. Crucially, it runs in real time across the full hardware spectrum, from a server GPU to a commodity smartphone CPU, within practical streaming-stride budgets.
\end{itemize}


\emph{By relocating the difficulty of LLM-based HAR from the language model to the sensor interface, \solution{} delivers a striking payoff: fed only sixteen compact tokens per window, a single \emph{frozen, unmodified} GPT-2 sets a new SOTA across diverse benchmark settings, outperforms pipelines built on LLMs orders of magnitude larger, runs in real time on a commodity smartphone, and personalizes to individual users without any data ever leaving the device. To the best of our knowledge, \solution{} is the first LLM-based HAR framework that is edge-deployable, configuration-invariant, and personalizable on-device at once, turning HAR into a lightweight, private modality that plugs into the shared local LLMs edge platforms already run, rather than yet another task-specific model to deploy and maintain.}

%% file: Files/Related_work.tex
\section{Related Work}

\subsection{Deep Learning for HAR}

Sensor-based HAR has long been studied as a time-series classification problem over wearable and mobile signals. Early deep learning approaches used CNNs to learn local temporal patterns from raw inertial data~\cite{zeng2014convolutional}, while recurrent models such as LSTMs and GRUs captured sequential dependencies across activity windows~\cite{guan2017ensembles, hammerla2016deep}, and hybrid architectures such as DeepConvLSTM combined the two~\cite{ordonez2016deep}. More recent systems adopt attention, transformer-style architectures, contrastive and metric learning, domain adaptation, and augmentation-based training to improve activity representations across users, devices, and sensor placements~\cite{khaertdinov2021deep, hermans2017defense, abedin2021attend, ahmad2023alae, zhao2020local, haresamudram2023investigating}. In parallel, Large Language Models (LLMs) have surpassed conventional deep models on many sequence understanding and reasoning tasks, motivating the HAR community to ask whether their semantic knowledge and sequence modelling can also benefit HAR.

\subsection{LLMs for HAR}

A growing body of work explores LLMs for HAR and related wearable sensing. The key motivation is semantic: activity labels such as walking upstairs or vacuuming are physical behaviours already represented in natural language, so an LLM can potentially connect low-level sensor evidence to high-level activity semantics. HARGPT~\cite{ji2024hargpt} directly prompts an LLM with sensor readings for zero-shot recognition, while other work uses learned activity embeddings~\cite{hota2025evaluating}, personalized wearable reasoning~\cite{khasentino2025personal}, and LLM-based contextualization of activity events~\cite{post2025contextllm, civitarese2025large}. SensorLLM~\cite{li2025sensorllm} aligns sensor signals with language-model representations through sensor-language alignment and task-aware tuning, LLM4HAR~\cite{hong2025llm4har} adapts pretrained LLMs for cross-domain HAR, and RAG-HAR~\cite{sivaroopan2025rag} uses retrieval-augmented prompting to supply similar activity examples for strong training-free performance. These studies establish LLMs as a promising direction for HAR. HAR ultimately targets wearable and mobile devices, however, raising a distinct challenge: how can LLM-based HAR be made efficient enough for edge devices without losing the benefits of pretrained language models?

\subsection{Edge-efficient Inference with LLMs for HAR}

There are two broad routes to practical LLM-based HAR. The first keeps the LLM off-device and queries it remotely via API. Work such as RAG-HAR~\cite{sivaroopan2025rag} takes this route: sensor-derived descriptions and retrieved examples are sent to a hosted LLM, which returns a prediction. This avoids local deployment but carries two costs for HAR. First, wearable data exposes sensitive information about gait, health, and routines, so transmitting it to the cloud raises privacy concerns. Second, even with a compact representation, network dependency and API latency are difficult to guarantee under the real-time constraints of continuous recognition.

The second route brings the LLM on-device, and can be approached in two ways. One rebuilds or compresses the LLM into a HAR-specific model. LLM4HAR~\cite{hong2025llm4har} starts from GPT-2, strips its text embedding and output-generation pathway, keeps selected decoder blocks, and adds sensor embeddings and an activity projection head before fine-tuning selected components. The result is compact but is no longer a general-purpose LLM, it is a GPT-2-derived classifier specialized for a fixed configuration (six-channel IMU, four shared activities), making it hard to extend to more channels, modalities, or classes, and unusable for other language or sensing tasks once its language I/O is removed. \solution{} does not pursue further compression; it asks instead whether a small pretrained LLM can remain a reusable on-device backend while the efficiency problem is solved at the sensor interface.

The other on-device approach keeps the language model largely intact and attaches a sensor tokenizer. SensorLLM~\cite{li2025sensorllm} pioneered this, using a pretrained Chronos~\cite{ansari2024chronos} encoder to produce sensor embeddings, projecting them into the LLM space, and inserting channel-specific tokens. This establishes a valuable sensor-to-LLM alignment paradigm but is not built for the edge: Chronos-large alone holds roughly 710M parameters atop an 8B-parameter LLM backbone, each channel is tokenized separately so token count grows with channel count and temporal segmentation, and the LLM-side pipeline is fine-tuned through sensor-language alignment and task-aware tuning rather than used as an unchanged service.

For edge inference, token count matters as much as model size: long sensor-derived sequences enlarge attention computation and the KV cache, increasing memory pressure, a problem amplified for HAR, where a window can hold hundreds to thousands of scalars across channels. \solution{} targets this directly. Rather than encoding channels separately or rebuilding the LLM, it learns a hierarchical tokenizer that compresses the entire multi-channel window into a compact set of latent tokens, which are projected into the embedding space and consumed by an unchanged pretrained LLM. This keeps the LLM reusable while making HAR inference predictable and efficient on-device, shifting the burden from LLM compression to sensor tokenization.

%% file: Files/Methodology-v2.tex
\begin{figure}
  \includegraphics[width=\columnwidth]{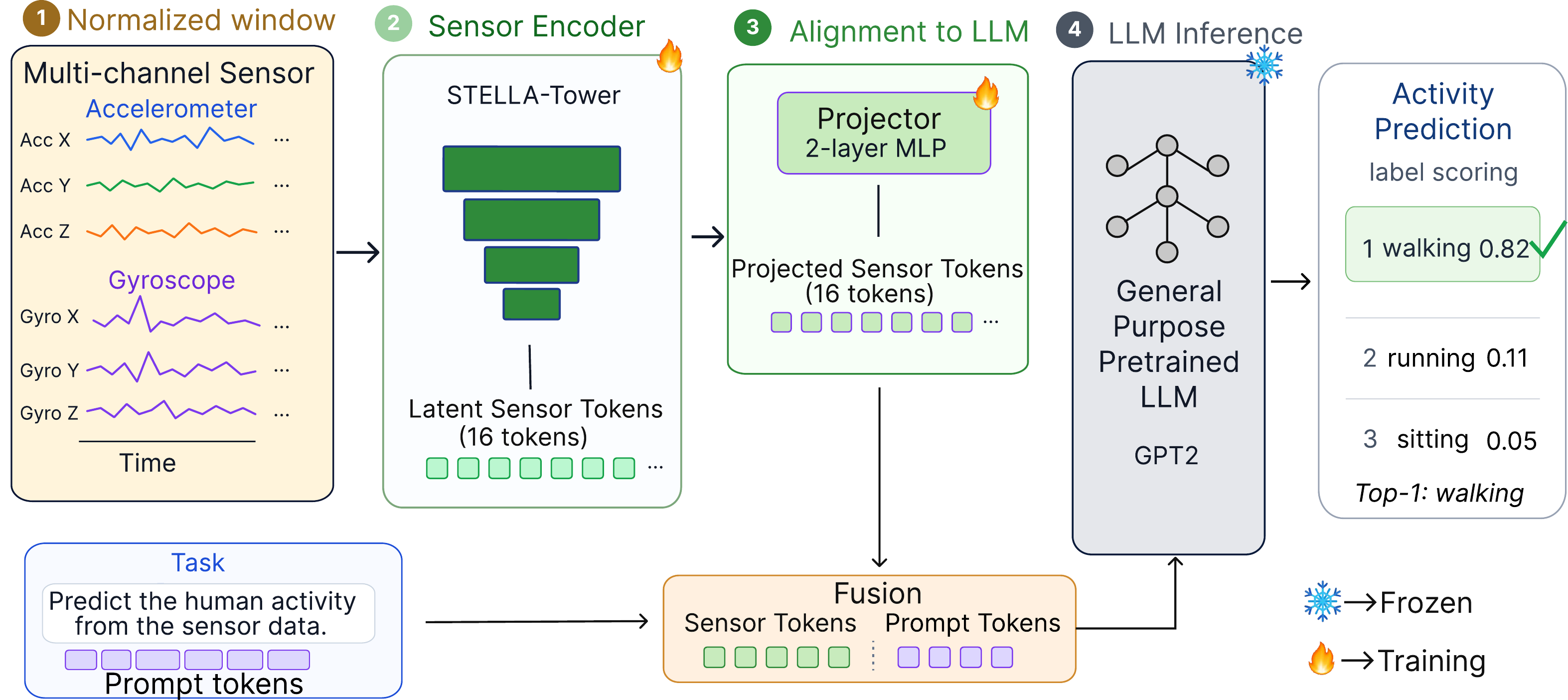}
  \vspace{-7mm}
  \caption{Overview of \solution{}. }
  \vspace{-6mm}
  \label{fig:overview}
\end{figure}

\begin{figure*}
  \includegraphics[width=\linewidth]{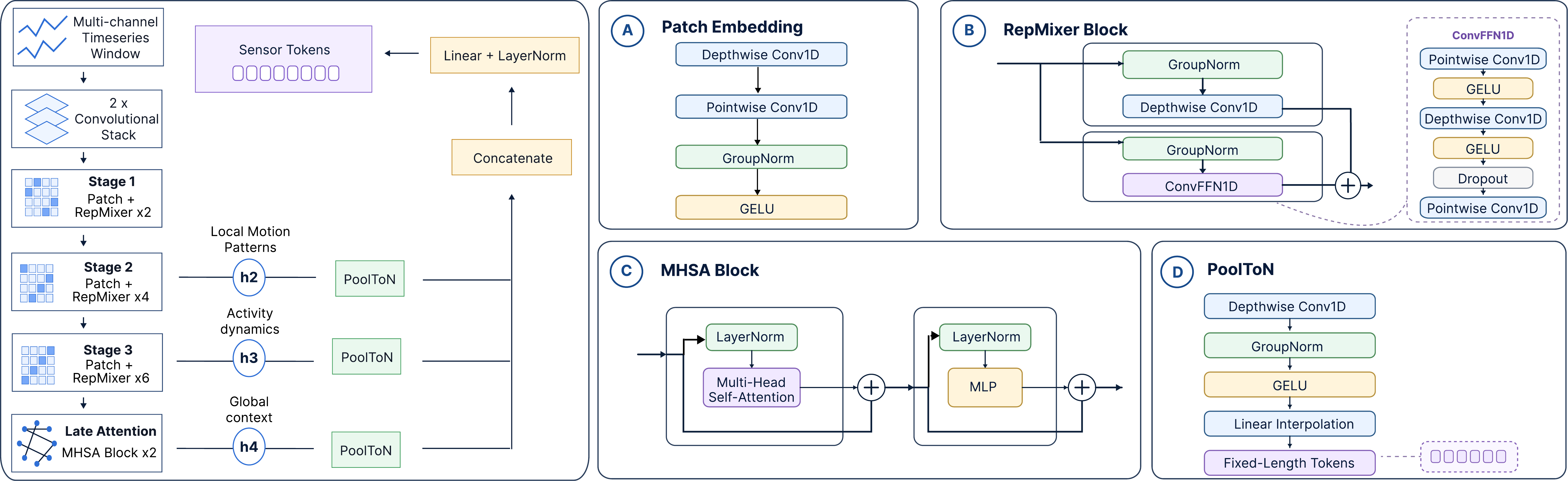}
   \vspace{-8mm}
  \caption{\solution{}-Tower. (a) Patch Embedding (b) RepMixer (c) MHSA (d) PoolToN blocks}
  \vspace{-4mm}
  \label{fig:Tokenizer}
\end{figure*}

\section{Methodology}

We propose \solution{}, a multimodal architecture that bridges continuous sensor signals and pretrained general-purpose LLMs through a learned tokenization interface for sensor stream reasoning. The key idea is to transform a raw multi-channel sensor window $\mathbf{X} \in \mathbb{R}^{C \times T}$, comprising $C$ sensor channels sampled over $T$ timesteps, into a short sequence of latent tokens that can be directly consumed by an LLM, enabling classification via language modeling. Unlike conventional pipelines that rely on task-specific classifiers, \solution{} formulates sensor understanding as a conditional text generation problem. This allows the model to leverage the rich representational structure of pretrained LLMs while maintaining an efficient and modular sensor encoder. As illustrated in Figure~\ref{fig:overview}, \solution{} encodes $\mathbf{X}$ into a fixed set of sensor tokens, projects them into the LLM embedding space, fuses them with a natural-language task prompt, and scores candidate activity labels with a frozen LLM.

\subsection{Design Overview}

The design of \solution{} is guided by three core principles:

\begin{itemize}
    \item \textit{Compatibility with LLMs:} Sensor representations must be mapped into the embedding space of pretrained LLMs without modifying their architecture.
    \item \textit{Efficiency for time-series data:} Sensor signals are long and high-frequency; the encoder must reduce temporal resolution while preserving salient patterns.
    \item \textit{Multi-scale representation:} Human activities exhibit patterns at different temporal scales, requiring hierarchical feature extraction.
\end{itemize}

To satisfy these requirements, \solution{} consists of three main components, which map directly onto the pipeline in Figure~\ref{fig:overview}:

\begin{enumerate}
    \item A \textit{sensor tokenizer}  that converts the raw window $\mathbf{X}$ into a fixed-length sequence of $N$ sensor tokens.
    \item A \textit{projection module} that aligns these sensor tokens with the LLM embedding space.
    \item A \textit{pretrained causal LLM} that performs prediction via next-token modeling.
\end{enumerate}

\subsection{Sensor Tokenizer}

The sensor tokenizer, shown as the \solution{}-Tower in Figure~\ref{fig:Tokenizer}, is the central component of \solution{}. It is a hierarchical encoder that progressively abstracts the raw time-series window into a compact set of semantic tokens.

\subsubsection{Motivation}

Raw sensor signals are: high-frequency (hundreds of timesteps), multi-channel (accelerometer, gyroscope, etc.), noisy and locally structured. Directly feeding such data into an LLM is infeasible due to sequence length and modality mismatch. The tokenizer addresses this by: reducing temporal resolution, increasing feature abstraction, and producing a small set of informative tokens.

\subsubsection{Convolutional Stack - Early Feature Extraction}

The tokenizer begins with a convolutional stack composed of depthwise separable convolutions. This design is chosen for efficiency and inductive bias:  (i) \textit{Depthwise convolution} extracts temporal patterns independently per channel, which is well-suited for sensor modalities where each channel captures distinct physical signals. (ii) \textit{Pointwise convolution} mixes information across channels, enabling cross-sensor interactions. The convolutional stack performs initial downsampling and maps the raw window $\mathbf{X}$ into a richer feature space:
\begin{equation}
\mathbf{H}_0 = f_{\text{conv\_stack}}(\mathbf{X}), \qquad \mathbf{H}_0 \in \mathbb{R}^{d_0 \times T_0},
\end{equation}
where $d_0$ is the feature dimension and $T_0 < T$ the reduced temporal length. This stage removes high-frequency noise while preserving coarse motion dynamics, and $\mathbf{H}_0$ serves as the input to the first hierarchical stage.

\subsubsection{Hierarchical Stages}

The encoder body consists of three stages indexed by $i \in \{1,2,3\}$, each comprising a \textit{patch embedding} layer followed by a stack of \textit{RepMixer blocks}. Stage $i$ transforms the feature map $\mathbf{H}_{i-1}$ into $\mathbf{H}_{i} \in \mathbb{R}^{d_i \times T_i}$, with progressively larger feature dimension $d_i$ and shorter temporal length $T_i$.

\paragraph{Patch Embedding.}
Each stage begins with a patch embedding layer (Figure~\ref{fig:Tokenizer}(a)) that performs temporal downsampling via strided depthwise--pointwise convolution. Conceptually, it groups adjacent timesteps into ``patches'' and projects them into a higher-dimensional feature space, reducing sequence length while increasing representational capacity:
\begin{equation}
\mathbf{H}_i = \mathrm{PatchEmbed}_i(\mathbf{H}_{i-1}), \qquad i \in \{1,2,3\}.
\end{equation}

\paragraph{RepMixer Blocks.}
Following patch embedding, each stage applies multiple \textit{RepMixer} blocks (Figure~\ref{fig:Tokenizer}(b)). The term ``RepMixer'' (short for \emph{reparameterizable mixer}) refers to a convolution-based alternative to self-attention that mixes information along the temporal dimension. A RepMixer block maps an input feature map $\mathbf{x}$ to an output $\mathbf{x}''$ through two residual sublayers:
\begin{align}
\mathbf{x}' &= \mathbf{x} + \mathrm{DWConv}(\mathrm{GN}(\mathbf{x})), \\
\mathbf{x}'' &= \mathbf{x}' + \mathrm{ConvFFN}(\mathrm{GN}(\mathbf{x}')),
\end{align}
where $\mathrm{GN}(\cdot)$ denotes group normalization, $\mathrm{DWConv}(\cdot)$ a depthwise convolution that performs \emph{temporal mixing}, and $\mathrm{ConvFFN}(\cdot)$ a convolutional feed-forward network that performs \emph{channel mixing}. The patch embedding together with the stacked RepMixer blocks of stage $i$ produce the stage output $\mathbf{H}_i$. We use RepMixer blocks instead of attention in these early stages for two reasons:

\begin{itemize}
    \item \textit{Efficiency:} Convolutions scale linearly with sequence length, making them suitable for long time-series and requiring less computation than attention.
    \item \textit{Local inductive bias:} Early sensor patterns are predominantly local (e.g., periodic motion).
\end{itemize}

\subsubsection{Late Attention Layer}

After the three convolutional stages, we apply a lightweight \textit{multi-head self-attention (MHSA)} module (Figure~\ref{fig:Tokenizer}(c)) to the stage-3 output $\mathbf{H}_3$. Because $\mathbf{H}_3$ has already been aggressively downsampled, this layer operates on a much shorter sequence than the raw window, making attention computationally feasible. Its role is to capture long-range dependencies, integrate information across the entire time window, and refine global context. The MHSA block follows a transformer structure, mapping its input $\mathbf{x}=\mathbf{H}_3$ to the attention-enhanced representation $\mathbf{H}_4 = \mathbf{x}''$:
\begin{align}
\mathbf{x}' &= \mathbf{x} + \mathrm{SelfAttention}(\mathrm{LayerNorm}(\mathbf{x})), \\
\mathbf{x}'' &= \mathbf{x}' + \mathrm{MLP}(\mathrm{LayerNorm}(\mathbf{x}')).
\end{align}
We place attention \emph{after} the convolutional stages to balance efficiency and expressiveness.

\subsubsection{Multi-scale Token Aggregation}

Sensor activities manifest at different temporal scales, so \solution{} aggregates features from three levels of the hierarchy, annotated in Figure~\ref{fig:Tokenizer} as local motion patterns, activity dynamics, and global context:

\begin{itemize}
    \item the intermediate representation $\mathbf{H}_2$ (stage-2 output),
    \item the high-level representation $\mathbf{H}_3$ (stage-3 output),
    \item the attention-enhanced representation $\mathbf{H}_4$.
\end{itemize}

Each selected representation $\mathbf{H}_i$ is passed through a \textit{PoolToN} module (Figure~\ref{fig:Tokenizer}(d)), which smooths features with a depthwise convolution, normalizes activations, and resamples the temporal axis to a fixed number of $N$ tokens via linear interpolation:
\begin{equation}
\mathbf{Z}_i = \mathrm{PoolToN}(\mathbf{H}_i), \qquad i \in \{2,3,4\}, \quad \mathbf{Z}_i \in \mathbb{R}^{N \times d_i}.
\end{equation}
Because every $\mathbf{Z}_i$ has the same token count $N$ regardless of its source resolution, the three representations are concatenated along the feature dimension and fused by a learned linear projection $\mathbf{W}$ followed by layer normalization:
\begin{equation}
\mathbf{S} = \mathrm{LayerNorm}(\mathbf{W} \cdot (\mathbf{Z}_2 \oplus \mathbf{Z}_3 \oplus \mathbf{Z}_4)), \qquad \mathbf{S} \in \mathbb{R}^{N \times d}.
\end{equation}
This yields the fixed-length sensor-token sequence $\mathbf{S}$ of $N$ tokens, capturing both local and global temporal patterns. Crucially, $N$ is fixed independently of $C$ and $T$, so downstream LLM cost does not grow with channel count or window length.

\subsection{Projection to LLM Space}

The sensor tokens $\mathbf{S}$ are mapped into the LLM embedding space using a two-layer MLP:
\begin{equation}
\mathbf{E}_s = f_{\text{proj}}(\mathbf{S}), \qquad \mathbf{E}_s \in \mathbb{R}^{N \times d_{\text{LLM}}},
\end{equation}
where $d_{\text{LLM}}$ is the token-embedding dimension of the pretrained LLM. This projection is necessary because the sensor tokenizer operates in a feature space of dimension $d$ that differs from the LLM space of dimension $d_{\text{LLM}}$; the MLP learns to align the two representations without modifying the LLM.

\subsection{Prompt Conditioning and LLM Integration}

In parallel, a textual prompt describing the task is embedded with the LLM's own tokenizer and embedding table:
\begin{equation}
\mathbf{E}_t = \mathrm{Embed}(\text{Prompt}), \qquad \mathbf{E}_t \in \mathbb{R}^{T_p \times d_{\text{LLM}}},
\end{equation}
where $T_p$ is the number of prompt tokens. The final input sequence is formed by concatenating the projected sensor embeddings and the prompt embeddings along the sequence dimension:
\begin{equation}
\mathbf{E} = [\mathbf{E}_s ; \mathbf{E}_t].
\end{equation}
Conditioned on $\mathbf{E}$, the LLM defines a distribution over candidate label token sequences,
\begin{equation}
P(y \mid \mathbf{X}, \text{Prompt}),
\label{eq:cond}
\end{equation}
which allows the model to interpret the sensor tokens within a natural-language context. This is the distribution that \solution{} maximizes during training and scores at inference.

\subsection{Training Objective}
\label{sec:training}
We train \solution{} with a standard autoregressive language-modeling objective. The target activity label is first tokenized by the LLM tokenizer into a sequence of $L$ tokens $y = (y_1, \ldots, y_L)$. The model is trained to predict this sequence under the conditional distribution of Eq.~\eqref{eq:cond}: each label token $y_t$ is predicted from the fused sensor--prompt input $\mathbf{E} = [\mathbf{E}_s ; \mathbf{E}_t]$ together with the label tokens that come before it, $y_1, \ldots, y_{t-1}$. The objective is the average negative log-likelihood over the $L$ label tokens:
\begin{equation}
\mathcal{L} \;=\; -\frac{1}{L} \sum_{t=1}^{L} \log P\!\left(y_t \,\middle|\, \mathbf{E},\, y_1, \ldots, y_{t-1}\right).
\end{equation}
The $1/L$ normalization makes the loss of each example independent of how many sub-tokens its label spans, so that multi-token labels (e.g., \emph{walking upstairs}) do not exert disproportionately large gradients compared to single-token labels (e.g., \emph{running}). In practice, the LLM is frozen, and only the sensor tokenizer and projection layers are trained. This significantly reduces training cost while leveraging pretrained knowledge.

\subsection{Inference}
\label{sec:inference}
At inference time, the candidate activity labels for the dataset form a closed set $\mathcal{Y}$, and we cast classification as label scoring: each candidate label is evaluated under the frozen LLM and the highest-scoring one is selected,
\begin{equation}
\hat{y} \;=\; \arg\max_{y \in \mathcal{Y}} \; s(y \mid \mathbf{X}, \text{Prompt}),
\end{equation}
where $\hat{y}$ is the predicted label and $s(\cdot)$ is the score of a candidate, defined as the average per-token log-likelihood the frozen LLM assigns to its token sequence. For a candidate label $y$ tokenized into $L$ tokens $(y_1, \ldots, y_L)$, each token is conditioned on the same fused input $\mathbf{E}$ and on the candidate's own preceding tokens:
\begin{equation}
s(y \mid \mathbf{X}, \text{Prompt}) \;=\; \frac{1}{L} \sum_{t=1}^{L} \log P\!\left(y_t \,\middle|\, \mathbf{E},\, y_1, \ldots, y_{t-1}\right).
\end{equation}
Because $L$ varies across candidate labels, the $1/L$ normalization, identical to the one used in training, keeps scores comparable between short and long label names.

\begin{figure*}
\centering
\includegraphics[width=\linewidth]{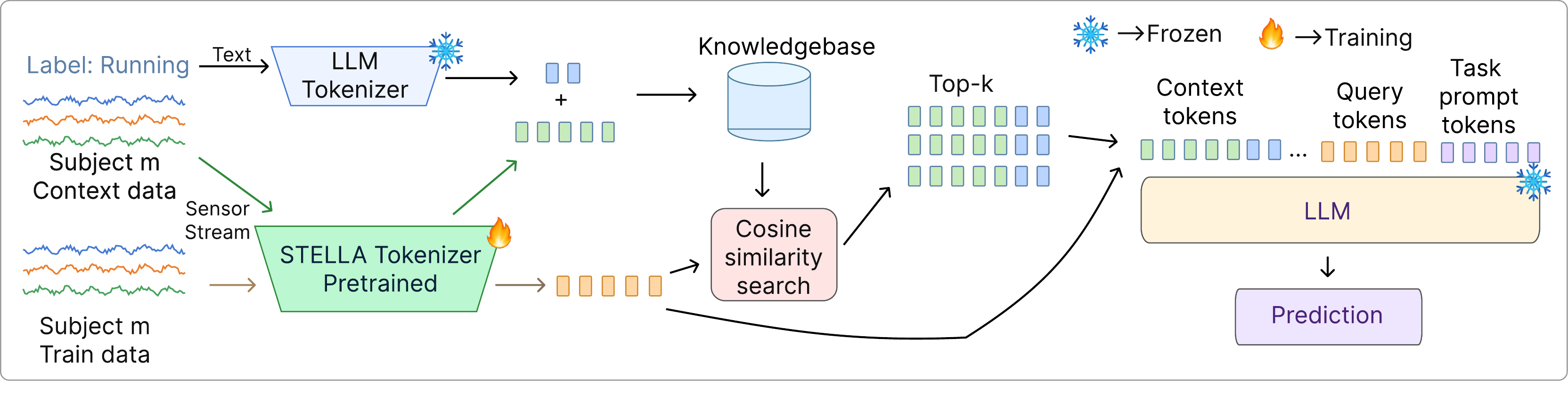}
\vspace{-10mm}
\caption{On-device personalization in \solution{}. }
\vspace{-6mm}
\label{fig:personalization}
\end{figure*}
 
\section{On-device Personalization}
\label{sec:ondevice_personalization}
 
\solution{} is designed not only for generic on-device HAR, but also for personalized recognition after deployment. Different users often perform the same activity with different motion patterns due to differences in gait, speed, body posture, device placement, and execution style. A model trained across many subjects provides a strong general representation, but can still benefit from adapting to the motion characteristics of a specific user. To support this, we introduce an on-device personalization mechanism that combines lightweight subject-specific tokenizer adaptation with retrieval-augmented inference. Figure~\ref{fig:personalization} summarizes the pipeline: the sensor tokenizer is the only component that is updated, while the LLM and its text tokenizer remain frozen throughout.
 
The personalization pipeline starts from a pretrained \solution{} model obtained from population-level training. After deployment, the device collects a small amount of labelled data from the target user, for example through lightweight calibration when a user selects an activity on a smartwatch and performs it briefly. As shown on the left of Figure~\ref{fig:personalization}, these samples are split into two roles: a \emph{training set} that adapts the lightweight sensor tokenizer, and a \emph{context set} stored in an on-device knowledge base for retrieval during inference. Updating only the sensor-side module preserves the LLM as a reusable on-device reasoning backend and avoids expensive full-model fine-tuning.
 
\paragraph{Tokenizer adaptation.}
 Starting from the pretrained tokenizer parameters, \solution{} fine-tunes the sensor tokenizer on training set using the same language-modeling objective as the main training stage (Section~\ref{sec:training}). This encourages the personalized tokenizer to produce sensor tokens that better align the user's motion patterns with the corresponding activity labels, while the frozen LLM continues to score candidate labels exactly as in standard inference (Section~\ref{sec:inference}). Consistent with Figure~\ref{fig:personalization}, gradients flow only through the tokenizer branch; the text tokenizer and LLM are never updated.
 
\paragraph{Knowledge base construction.}
The knowledge base is populated from the context set. For each context window, the current sensor tokenizer produces a sensor-token embedding, and the frozen LLM tokenizer converts the associated activity label into label tokens; the two are stored together, so each entry pairs a personalized sensor embedding with its label. Because the tokenizer changes during adaptation, the knowledge base is refreshed after every personalization epoch: all context samples are re-encoded with the updated tokenizer and their stored embeddings replaced. This keeps retrieval consistent with the current personalized representation space rather than relying on stale embeddings from an earlier tokenizer state.
 
\paragraph{Retrieval-augmented inference.}
At inference, a test window is first encoded by the personalized tokenizer into a query embedding. \solution{} then runs a cosine-similarity search against each stored context embedding in the knowledge base, and retrieves the top-$k$ most similar context examples as user-specific evidence. Following the figure, the final LLM input prepends the retrieved context tokens $\mathbf{E}_{\mathrm{ctx}}$ to the query sensor tokens $\mathbf{E}_s$ and the task-prompt tokens $\mathbf{E}_t$, extending the standard fused input of Eq.~\eqref{eq:cond} to $\mathbf{E}_{\mathrm{pers}} = [\mathbf{E}_{\mathrm{ctx}} ; \mathbf{E}_s ; \mathbf{E}_t]$. The frozen LLM then predicts the activity by scoring candidate label sequences over $\mathbf{E}_{\mathrm{pers}}$, exactly as in the standard \solution{} inference procedure (Section~\ref{sec:inference}).
 
This design has two advantages for on-device HAR. First, personalization is lightweight: adaptation is restricted to the compact sensor tokenizer rather than the full LLM. Second, retrieval is local and privacy-preserving, since the user's sensor data and context memory never leave the device. 

%% file: Files/Results-v4.tex
\section{Results}

We evaluate \solution{} on seven widely used public HAR benchmarks: HHAR~\cite{stisen2015smart}, PAMAP2~\cite{reiss2012introducing}, MHEALTH~\cite{banos2014mhealthdroid}, GOTOV~\cite{paraschiakos2020activity}, SKODA~\cite{stiefmeier2008wearable}, Opportunity~\cite{roggen2010collecting} and USC-HAD~\cite{zhang2012usc}, which span 6 to 60 sensor channels and 6 to 18 activities, covering diverse activity types such as locomotion (e.g., walking, running), daily living tasks (e.g., ironing, vacuuming), fitness exercises (e.g., cycling, jogging), manufacturing assembly-line tasks, and settings that capture activity variations across pace, direction, or equipment. To ensure direct comparability and consistency with prior peer-reviewed benchmarks (Percom, SenSys, EMNLP), we replicate dataset-specific preprocessing and evaluation protocols (train-test split) from earlier works: HHAR follows~\cite{haresamudram2023investigating}, PAMAP2 and Opportunity (gestures) follow~\cite{jeyakumar2019sensehar}, Opportunity (locomotion) follows~\cite{li2021units}, MHEALTH follows~\cite{suh2022adversarial}, GOTOV and SKODA follow~\cite{ahmad2023alae}, and USC-HAD follows~\cite{khaertdinov2021deep}. \footnote{per-dataset pre-processing and evaluation protocols are detailed in Appendix~\ref{app:dataset}}  
Sections~\ref{sec:har-benchamark} and~\ref{sec:tokenizer} benchmark \solution{} against prior HAR methods and time-series tokenizers, Section~\ref{sec:feasibility} analyzes on-device feasibility, and Section~\ref{sec:ablation} ablates the design. The personalization step of Section~\ref{sec:ondevice_personalization} is applied only in Section~\ref{sec:personalization}.

\subsection{HAR Benchmark}
\label{sec:har-benchamark}

\input{Files/main-table}

Table~\ref{tab:results-comparison-table} reports \solution{} against prior deep learning and LLM-based approaches across all eight benchmarks. \solution{} achieves the best reported performance among the compared methods on every benchmark under the corresponding published protocol, with gains ranging from marginal on near-saturated benchmarks to substantial on sparse-sensor platforms.

\paragraph{Low-channel benchmarks}
The largest gains occur on the two single-IMU datasets, where sensor evidence is most constrained. On USC-HAD, \solution{} reaches 72.00\% F1, +9.20 over the best prior baseline (Triplet LSTM HTL-SB) and +10.80 over the strongest prior LLM method (SensorLLM). On HHAR it reaches 71.69\% F1, +11.83 over the best prior method (RAG-HAR) and +12.44 over the best contrastive baseline (Enhanced CPC). 
RAG-HAR, which classifies by retrieving similar labelled segments, is especially disadvantaged here, since 6-channel representations are less distinctive for retrieval while \solution{} achieves larger metrics even with less number of channels.

\paragraph{Rich-sensor and near-saturated benchmarks}
\solution{} also leads where prior methods already perform strongly, confirming it does not sacrifice recognition performance for efficiency. On MHEALTH it reaches 97.15\% F1, the highest reported, exceeding RAG-HAR by +0.41 and ADFE by +0.68. On Skoda it reaches 95.87\% F1, beating RAG-HAR by +0.66 and ALAE-TAE-CutMix+ by +1.07. On Opportunity it reaches 94.29\% F1 (locomotion), +0.96 over the prior SenSys best Uni-TS, and 69.75\% F1 (gesture), +2.27 over SenseHAR. On GOTOV it reaches 82.60\% F1, +2.68 over RAG-HAR and +3.20 over ALAE-TAE-CutMix+. On PAMAP2 it sets a new SOTA at 96.52\% F1, +1.44 over SenseHAR.

\paragraph{Comparison with LLM-based approaches.}
The LLM-based baselines fall into two lines, both limited for on-device use. The API line prompts a cloud LLM: HARGPT~\cite{ji2024hargpt} feeds the raw window directly and scales poorly as channels and sampling rate grow, while LLM as VA~\cite{hota2025evaluating} instead passes a two-value t-SNE summary that is too compressed to capture multi-channel dynamics, both score poorly in Table~\ref{tab:results-comparison-table}. RAG-HAR~\cite{sivaroopan2025rag} is the strongest prior method, using per-channel statistical descriptors and retrieved examples, but its prompt grows with channels and context, costing more than 6\,s per prediction on a 6-channel dataset and requiring sensor data to be sent to the cloud. The fine-tuning line, represented by SensorLLM, tokenizes each channel separately with Amazon's Chronos~\cite{ansari2024chronos} encoder (roughly 710M parameters) atop an 8B-parameter LLM, so both token count and model size scale unfavourably for the edge.

\solution{} avoids all of these. It compresses each complete window into a fixed set of 16 latent tokens before any LLM computation, a 262.5$\times$ reduction for MHEALTH and 187.5$\times$ for USC-HAD over scalar-level tokenization, so inference cost is independent of window length, and it keeps the LLM frozen with the full pipeline local,  so sensor data need not leave the device. Inference completes in under 10\,ms on a server GPU and under 50\,ms on a mobile GPU (Section~\ref{sec:feasibility}), versus multi-second API latency for RAG-HAR. Despite using only a frozen GPT-2 backbone, \solution{} outperforms every prior LLM-based method where a direct comparison exists, exceeding RAG-HAR by 0.41--13.37 F1 points (largest on USC-HAD and HHAR). The key to LLM-based HAR is therefore not a larger LLM, a longer prompt, or heavier fine-tuning, but an efficient sensor-to-LLM translation layer that preserves activity-relevant information while keeping the LLM input short, private, and deployable on-device.

\subsection{Tokenizer Performance}
\label{sec:tokenizer}

\begin{table}[]
\centering
\caption{Benchmarking Tokenizer performance}
\vspace{-4mm}
\label{tab:tokenizer_performance}
\small
\setlength{\tabcolsep}{3.5pt}
\begin{tabular}{crrrrrrr}
\toprule
& \multicolumn{2}{c}{\textbf{HHAR}} &
  \multicolumn{2}{c}{\textbf{MHEALTH}} &
  \multicolumn{2}{c}{\textbf{USC-HAD}} \\
\cmidrule(lr){2-3}\cmidrule(lr){4-5}\cmidrule(lr){6-7}
Tokenizers &
  \makecell{\textbf{Acc}\\\textbf{(\%)}} &
  \makecell{\textbf{F1}\\\textbf{(\%)}} &
  \makecell{\textbf{Acc.}\\\textbf{(\%)}} &
  \makecell{\textbf{F1}\\\textbf{(\%)}} &
  \makecell{\textbf{Acc.}\\\textbf{(\%)}} &
  \makecell{\textbf{F1}\\\textbf{(\%)}} \\
\midrule
\textbf{\solution{}-Tower (ours)}         & \textbf{72.71}   & \textbf{71.69}  & \textbf{96.68} & \textbf{96.22} & \textbf{86.92} &\textbf{81.01} \\
PatchTST~\cite{nie2022time}                     & 64.74   & 61.73  & 94.52 & 94.21 & 78.93 & 74.38 \\
Chronos~\cite{ansari2024chronos}                & 62.46   & 60.73  & 92.78 & 92.97 & 79.39 & 75.06 \\
Timesnet~\cite{wu2022timesnet}                  & 61.06  & 59.64  & 94.71 & 94.30 & 76.55 & 70.92 \\
itransformer~\cite{liu2024itransformer}         & 65.21  & 64.03  & 88.52 & 87.26 & 79.02 & 74.39 \\
Tsmixer~\cite{ekambaram2023tsmixer}             & 54.11  & 51.66  & 94.64 & 94.22 & 73.86 & 69.04 \\
Timemixer~\cite{wang2024timemixer}              & 59.21  & 58.76  & 94.19 & 93.50 & 80.90 & 75.94 \\
\bottomrule
\end{tabular}
\vspace{-4mm}
\end{table}

The central hypothesis of \solution{} is that LLM-based HAR should not rely on raw numerical prompts or generic time-series representations, but instead requires a sensor tokenizer that is explicitly designed to translate wearable signals into a compact sequence of LLM-compatible tokens. To evaluate this hypothesis, we compare our \solution{}-Tower tokenizer against representative SOTA time-series encoders, including PatchTST~\cite{nie2022time}, Chronos~\cite{ansari2024chronos}, TimesNet~\cite{wu2022timesnet}, iTransformer~\cite{liu2024itransformer}, TSMixer~\cite{ekambaram2023tsmixer}, and TimeMixer~\cite{wang2024timemixer} in Table~\ref{tab:tokenizer_performance}. For a fair comparison, all tokenizers are constrained to output the same number of  tokens, $N=16$, and are evaluated under the same downstream LLM-based classification pipeline. Thus, the comparison isolates the quality of the sensor-to-token translation rather than differences in LLM capacity or token budget. We perform a 5-fold held-out evaluation protocol for this experiment. The detailed implementation of the tokenizer baselines are presented in Appendix~\ref{app:tokenizer_details}. Table~\ref{tab:tokenizer_performance} shows that \solution{}-Tower consistently outperforms all competing tokenizers across HHAR, MHEALTH, and USC-HAD. On HHAR, \solution{}-Tower achieves 71.69\% F1, improving over the strongest baseline by +7.66 F1 points. 
On MHEALTH, where most encoders already perform strongly, \solution{}-Tower still achieves the best result with 96.22\% F1, surpassing the next best tokenizer by +1.92 F1 points. On USC-HAD, \solution{}-Tower reaches 86.92\% accuracy and 81.01\% F1, outperforming the strongest baseline by +5.07 F1 points. These results highlight an important distinction between general-purpose time-series encoders and sensor tokenizers for LLM-based HAR. Models such as PatchTST, TimesNet, iTransformer, TSMixer, and TimeMixer were primarily designed to model temporal dependencies in conventional time-series forecasting or classification settings. 
These encoders often lose fine-grained activity cues or fail to preserve the multi-scale structure needed for HAR. 
\solution{}-Tower is better suited to this bottleneck because its architecture is explicitly designed around compact sensor-to-LLM translation. 
Overall, Table~\ref{tab:tokenizer_performance} confirms that the performance of \solution{} is not only due to the downstream LLM, but also to the quality of the learned sensor tokenizer. 
These results validate the design choice of using a dedicated sensor tokenizer rather than directly adopting existing time-series encoders for LLM-based HAR.

\subsection{On-Device Feasibility Analysis}
\label{sec:feasibility}

We systematically characterise the computational footprint, inference latency, memory requirements, and deployment constraints of \solution{} to establish its viability for real-time, on-device HAR\@. To assess generalisation across sensor configurations, all experiments are conducted on two datasets with contrasting characteristics: MHEALTH (21 channels, 50\,Hz, 4-second windows) and USC-HAD (6 channels, 100\,Hz, 5-second windows). This pairing deliberately spans the extremes of practical wearable deployments, a rich multi-placement inertial suite versus a minimal single-IMU platform. Unless stated otherwise, latency figures report the median (p50) of 200 isolated benchmark iterations preceded by 50 warm-up rounds, using CUDA events for GPU timing and high-resolution wall-clock measurements for CPU\@. In a streaming HAR deployment, the sliding stride determines how frequently a new window arrives for classification; the system must complete each inference before the next window is ready. We therefore define the \textit{inference budget} as the stride duration: 1{,}000\,ms for MHEALTH (50-sample stride at 50\,Hz) and 2{,}500\,ms for USC-HAD (250-sample stride at 100\,Hz). A system is real-time feasible if inference latency\,<\,50\% of this budget, preserving headroom for sensor I/O and application overhead. We perform a 5-fold held-out evaluation protocol for these experiments.

\subsubsection{\textbf{Computational Profile}}
\label{sec:profile}

Table~\ref{tab:profile} summarises the parameter distribution and floating-point operations of \solution{} across both sensor configurations.
The model architecture is identical for both datasets; only the input dimension differs. \solution{} comprises 138.78\,M parameters, of which
117\,M (89.7\%) reside in the frozen GPT-2 backbone. The trainable component, sensor encoder plus MLP projector, totals 14.34\,M parameters
(10.3\%), confining gradient computation entirely to the lightweight sensor front-end.

\begin{table}[t]
\centering
\caption{Computational profile of \solution{}. GFLOPs are reported for a single forward pass through each component. A complete classification requires one LLM pass for each candidate activity label.}
\vspace{-4mm}
\label{tab:profile}
\small
\setlength{\tabcolsep}{5pt}
\begin{tabular}{lrrrr}
\toprule
\textbf{Component} & \textbf{Params} & \textbf{Share} &
  \makecell{\textbf{GFLOPs}\\\textbf{(single pass)}} &
  \textbf{FP32 Size} \\
\midrule
Sensor Encoder  & 12.57\,M  &  9.1\% & 0.12  & 50.3\,MB  \\
MLP Projector   &  1.77\,M  &  1.3\% & $<$0.01 & 7.1\,MB \\
GPT-2 (frozen)  & 117\,M & 89.7\% & 7.42  & 497.8\,MB \\
\midrule
\textbf{Total}  & \textbf{138.78\,M} & \textbf{100\%} &   \textbf{7.54} & \textbf{555.2\,MB} \\
\bottomrule
\end{tabular}
\vspace{-4mm}
\end{table}

The encoder contributes 1.6\% of the per-pass GFLOPs (0.12 of 7.54), confirming that the LLM is the dominant computational component regardless
of sensor configuration. Since classification scores all candidate labels, let $C = |\mathcal{Y}|$ denote the number of candidate activity labels. For MHEALTH, $C=12$, so the total inference cost per prediction is $C \times 7.42 + 0.12 \approx 89.2$ GFLOPs, the encoder runs once, and the LLM runs once per candidate label. The encoder FLOPs differ by 50\% across datasets (0.12 vs.\ 0.18\,GFLOPs) owing to USC-HAD's longer 500-sample windows; however, total inference latency FLOPs are virtually identical (89.10 vs.\ 89.16\,GFLOPs), because encoder computation accounts for just 0.13--0.20\% of the total budget. The LLM cost of 88.98\,GFLOPs is completely invariant to input channel count and window length, as it always processes the same 16-token compressed representation. Raw sensor input for MHEALTH spans 4{,}200 scalar values per window (200$\times$21), compressed to 16 tokens, a \textit{262.5$\times$ reduction} relative to na\"ive per-scalar tokenisation. USC-HAD spans 3{,}000 scalar values (500$\times$6), compressed to the same 16 tokens, a \textit{187.5$\times$ reduction}.

\subsubsection{\textbf{Inference Latency}}
\label{sec:latency}

\begin{table}[t]
\centering
\caption{Latency breakdown (median of 200 runs, batch $= 1$, FP32). Budget: inference latency as a fraction of the prediction stride (1{,}000\,ms for MHEALTH; 2{,}500\,ms for USC-HAD).}
\vspace{-4mm}
\label{tab:latency}
\small
\setlength{\tabcolsep}{4pt}
\begin{tabular}{lllrrr}
\toprule
\textbf{DS} & \textbf{Device} & \textbf{Config} &
  \makecell{\textbf{Tokenizer}\\\textbf{(ms)}} &
  \makecell{\textbf{Inference}\\\textbf{Latency (ms)}} &
  \makecell{\textbf{Budget}\\\textbf{(\%)}} \\
\midrule
MH  & RTX 4090     & GPU FP32 &  1.62 &   9.73 &  0.97 \\
MH  & Server (x86) & CPU, 16T &  4.05 & 206.4  & 20.6  \\
MH  & Server (x86) & CPU,  4T &  4.01 & 247.8  & 24.8  \\
MH  & Server (x86) & CPU,  1T &  6.84 & 852.2  & 85.2  \\
\midrule
USC & RTX 4090     & GPU FP32 &  1.53 &   9.74 &  0.39 \\
USC & Server (x86) & CPU, 16T &  5.16 & 180.3  &  7.2  \\
USC & Server (x86) & CPU,  4T &  5.26 & 225.2  &  9.0  \\
USC & Server (x86) & CPU,  1T &  7.86 & 774.7  & 31.0  \\
\bottomrule
\end{tabular}
\vspace{-6mm}
\end{table}

We define \textit{inference latency} as the elapsed time from receiving a raw sensor window to producing a classification label, encompassing
sensor encoding and all $C$ LLM scoring passes. Table~\ref{tab:latency} presents this latency alongside its encoder sub-component across all
evaluated hardware configurations on the server with different cores. The central result is the near-perfect alignment of inference latency across the two sensor configurations. On GPU, MHEALTH and USC-HAD differ by \textit{0.01\,ms} (9.73 vs.\ 9.74\,ms) despite USC-HAD's 2.5$\times$
longer input window. On CPU (4 threads), USC-HAD is marginally \emph{faster} than MHEALTH (225.2 vs.\ 247.8\,ms), because the 3.5$\times$
reduction in channel count more than compensates for the 2.5$\times$ increase in temporal samples in the early depthwise encoder stages.
This confirms that end-to-end prediction latency is governed by the LLM processing 16 fixed tokens rather than by raw sensor volume, channel
count, or sampling rate.

Two further observations hold across both datasets. First, the LLM dominates CPU cost: the encoder contributes just 1.6--2.3\% of inference latency at 4~threads (4.01/247.8\,ms for MHEALTH; 5.26/225.2\,ms for USC-HAD). Second, USC-HAD's longer 2.5-second stride substantially relaxes the real-time constraint: even single-thread CPU inference (774.7\,ms, 31.0\% of the 2.5-second budget) is feasible, whereas on MHEALTH single-thread inference (852.2\,ms, 85.2\% budget) is not.

\subsubsection{\textbf{Memory Requirements}}
\label{sec:memory}

Peak GPU VRAM during a full $C$ batched prediction is 749.9\,MB on MHEALTH and 788.3\,MB on USC-HAD. The 38\,MB difference is attributable to
the larger intermediate activation tensors produced by USC-HAD's 500-sample input windows. Both values fit within the 1\,GB VRAM available on entry-level mobile GPUs (Adreno~619 and above). For CPU-only deployment, a minimum of \textit{4\,GB total device RAM} is required on both datasets, with the inference process consuming approximately 1.7\,GB (model weights 555\,MB, PyTorch runtime $\approx$900\,MB, and inference activations $\approx$183--221\,MB).

\subsubsection{\textbf{Precision and Quantisation}}
\label{sec:quant}


\begin{table}[t]
\centering
\caption{Effect of model compression on size, latency, and accuracy. $\dagger$~FP16 on x86 CPU emulates via FP32 upcasting; not representative of ARM CPUs with native FP16 arithmetic. $\ddagger$~LLM-only schemes quantize GPT-2 transformer block weights only; the embedding table and sensor encoder remain in FP32}
\vspace{-4mm}
\label{tab:quant}
\small
\setlength{\tabcolsep}{1.5pt}
\begin{tabular}{llcccccccc}
\toprule
& \textbf{Scheme} &
  \makecell{\textbf{Size}\\\textbf{(MB)}} &
  \makecell{\textbf{CPU 4T}\\\textbf{p50 (ms)}} &
  \makecell{\textbf{GPU}\\\textbf{p50 (ms)}} &
  \multicolumn{2}{c}{\textbf{MHEALTH}} &
  \multicolumn{2}{c}{\textbf{USC-HAD}} \\
\cmidrule(lr){6-7}\cmidrule(lr){8-9}
& & & & &
  \makecell{\textbf{Acc.}\\\textbf{(\%)}} &
  \makecell{\textbf{F1}\\\textbf{(\%)}} &
  \makecell{\textbf{Acc.}\\\textbf{(\%)}} &
  \makecell{\textbf{F1}\\\textbf{(\%)}} \\
\midrule
\multirow{3}{*}{\rotatebox{90}{\scriptsize \makecell{Full \\ Model}}}
& FP32 (baseline) & 555.2 & 247.8 &  9.73 & 96.68 & 96.22 & 86.92 & 81.01 \\
& FP16            & 277.7 & 2{,}146$^\dagger$ &  6.30 & 96.68 & 96.22 & 86.60 & 80.92 \\
& INT8            & 152.3 & 184.3 &  4.14 & 40.13 & 27.08 & 86.74 & 80.75 \\
\midrule
\multirow{2}{*}{\rotatebox{90}{\scriptsize \makecell{LLM\\ Only}}}
& INT8$^\ddagger$ & 300.8 & 252.2 &  8.91 & 96.58 & 96.48 & 86.59 & 80.68 \\
& INT4$^\ddagger$ & 263.3 & 237.8 &  6.34 & 96.52 & 96.41 & 85.93 & 78.02 \\
\bottomrule
\end{tabular}
\vspace{-4mm}
\end{table}

Table~\ref{tab:quant} evaluates two compression strategies: full-model compression, which quantises all layers including the GPT-2 embedding
table, and LLM-only compression, which quantises only the transformer block weights while keeping the embedding table and sensor encoder in FP32.

Full-model FP16 halves model size to 277.7\,MB and delivers a 35.3\% GPU speedup (6.30\,ms) with no accuracy loss on either dataset. On x86 CPU, FP16 is unusable due to the absence of hardware FP16 arithmetic; ARM processors with native FP16 units are not affected. Full-model INT8 produces the most compact checkpoint (152.3\,MB) and the fastest CPU inference (184.3\,ms). Its behaviour, however, is strongly dataset-dependent: accuracy is fully preserved on USC-HAD  but drops to 40.13\% on MHEALTH. The divergence arises from the sensitivity of NLL-based label scoring to quantisation-induced perturbations in the embedding table. On MHEALTH, where the model achieves near-perfect discrimination, inter-class NLL margins are narrow; quantising the embedding table distorts token-level log-probabilities enough to corrupt the relative label ordering. On USC-HAD, where activities are harder to distinguish and NLL margins are wider, the same perturbation stays below the decision boundary and accuracy is preserved.

LLM-only compression (HuggingFace) avoids embedding-table quantisation and recovers accuracy on both datasets. INT8 LLM-only reduces MHEALTH accuracy by 0.10 percentage points (96.58\%) and USC-HAD by 0.33 points (86.59\%). Its model size (300.8\,MB) exceeds FP16 because the unquantised embedding table ($\approx$154\,MB) dominates the footprint. On GPU, inference latency is 8.91\,ms; on CPU it is 252.2\,ms, marginally slower than FP32 (247.8\,ms) due to dequantisation overhead at transformer block boundaries. INT4 LLM-only yields the smallest accuracy-preserving checkpoint at \textit{263.3\,MB}, 5.2\% smaller than FP16 and 52.6\% smaller than FP32, while matching FP16 GPU speed (6.34\,ms, 34.8\% faster than FP32). On CPU it is 237.8\,ms, a 4.0\% improvement over FP32; the modest CPU gain reflects that compute, not weight loading, is the bottleneck at this sequence length. Accuracy costs are 0.16 percentage points on MHEALTH (96.52\%) and 0.99 points on USC-HAD (85.93\%).

For GPU deployment, INT4 LLM-only is preferable to FP16: it achieves equivalent latency (6.34\,ms vs.\ 6.30\,ms), uses less memory (263\,MB vs.\ 278\,MB), and incurs less than 0.2 percentage points accuracy loss on MHEALTH. For CPU deployment, FP32 remains the safe default; INT4 LLM-only is a viable alternative where storage matters, at the cost of a modest 4.0\% latency improvement. Full-model INT8 is appropriate only for datasets where the model operates below $\approx$90\% accuracy and NLL margins are correspondingly wide.

\subsubsection{\textbf{Effect of Sensor Token Count}}
\label{sec:tokens}

\begin{table}[]
\centering
\caption{Classification accuracy and inference latency vs.\ number of sensor tokens $N$.}
\vspace{-4mm}
\label{tab:tokens}
\small
\setlength{\tabcolsep}{3.5pt}
\begin{tabular}{crrrrrrr}
\toprule
& \multicolumn{2}{c}{\textbf{Latency (ms)}} &
  \multicolumn{2}{c}{\textbf{MHEALTH}} &
  \multicolumn{2}{c}{\textbf{USC-HAD}} \\
\cmidrule(lr){2-3}\cmidrule(lr){4-5}\cmidrule(lr){6-7}
$N$ &
  \makecell{\textbf{GPU}\\\textbf{p50}} &
  \makecell{\textbf{CPU 4T}\\\textbf{p50}} &
  \makecell{\textbf{Acc.}\\\textbf{(\%)}} &
  \makecell{\textbf{F1}\\\textbf{(\%)}} &
  \makecell{\textbf{Acc.}\\\textbf{(\%)}} &
  \makecell{\textbf{F1}\\\textbf{(\%)}} \\
\midrule
4           & 9.5   & 149  & 95.93 & 95.29 & 86.98 & 80.04 \\
8           & 9.9   & 164  & 96.13 & 95.68 & 87.18 & 80.90 \\
\textbf{16} & 10.1  & 211  & \textbf{96.68} & \textbf{96.22} & 86.92 & 81.01 \\
\textbf{32} & 10.7  & 301  & 96.12 & 95.40 & \textbf{87.45} & \textbf{81.58} \\
64          & 13.4  & 499  & 95.65 & 95.03 & 86.25 & 79.35 \\
\bottomrule
\end{tabular}
\vspace{-6mm}
\end{table}

Table~\ref{tab:tokens} reports accuracy and inference latency as the number of sensor tokens $N$ varies from 4 to 64 for both datasets. The optimal token count is dataset-dependent and correlates with window length. On MHEALTH (200-sample windows), $N{=}16$ is optimal; on USC-HAD
(500-sample windows), $N{=}32$ performs best, reflecting the higher information density of the longer temporal context. The compression ratios
at the respective optima are comparable: 12.5$\times$ temporal compression for MHEALTH (200\,samples\,$\to$\, 16\,tokens) and 15.6$\times$ for USC-HAD (500\,samples\,$\to$\,32\,tokens). Notably, $N{=}4$, a 50$\times$ compression from MHEALTH's 200-sample window and a 125$\times$ compression
from USC-HAD's 500-sample window, retains 95.93\% and 86.98\% accuracy respectively, demonstrating that the encoder extracts a highly compact
representation with minimal loss. On GPU, latency is nearly insensitive to $N$ across the 4--32 range (9.5--10.7\,ms) on both datasets, with
a step at $N{=}64$ (13.4\,ms) marking the onset of attention quadratic scaling. On CPU, the relationship is monotone, and $N{=}64$ reaches
the real-time threshold (499\,ms) on both datasets.

\subsubsection{\textbf{Real-Time Deployment Feasibility}}
\label{sec:deployment}


\begin{table*}[t]
\centering
\caption{Real-time deployment feasibility of \solution{} FP32 (batch $= 1$). Budget: inference latency as a fraction of the prediction stride (1{,}000\,ms for MHEALTH; 2{,}500\,ms for USC-HAD).
}
\vspace{-4mm}
\label{tab:deployment}
\small
\resizebox{\textwidth}{!}{%
\setlength{\tabcolsep}{5pt}
\begin{tabular}{llllrrrr}
\toprule
\textbf{Tier} & \textbf{Device} & \textbf{Cores / Accelerator} & \textbf{RAM} &
  \makecell{\textbf{Runtime}} &
  \makecell{\textbf{Latency}\\\textbf{p50 (ms)}} &
  \makecell{\textbf{Budget}\\\textbf{MH / USC}} &
  \makecell{\textbf{RT?}\\\textbf{MH / USC}}  \\
\midrule
\multirow{3}{*}{Server}
  & RTX 4090 (GPU) & 16,384 CUDA cores & 24\,GB VRAM & PyTorch CUDA      &   9.73  &  0.97\% /  0.39\%  & \checkmark / \checkmark \\
  & x86-64 server & 16$\times$ x86-64 & 134\,GB & PyTorch CPU        & 206.4   & 20.6\%  /  8.3\%   & \checkmark / \checkmark \\
  & x86-64 server & 4$\times$ x86-64  & 134\,GB & PyTorch CPU        & 247.8   & 24.8\%  /  9.9\%   & \checkmark / \checkmark  \\
\midrule
\multirow{1}{*}{Edge AI}
  & Jetson Xavier NX & 384-core Volta GPU & 8\,GB  & TensorRT    & 35.2 & 3.52\% / 1.4\% & \checkmark / \checkmark  \\
\midrule
\multirow{3}{*}{\makecell[l]{Smartphones}}
  & Samsung S23 Ultra & 8-core Snap.\ 8 Gen 2 & 12\,GB & ONNX CPU EP     &  377.8    &  37.8\%  /  15.1\%  & \checkmark / \checkmark \\
  & Samsung S23 Ultra & Adreno 740 GPU    & 12\,GB & ONNX NNAPI EP   & 41.7    & 4.17\% / 1.67\% & \checkmark / \checkmark  \\
\bottomrule
\end{tabular}}
\vspace{-4mm}
\end{table*}

Table~\ref{tab:deployment} consolidates the real-time feasibility of \solution{} across three deployment tiers: server-class hardware, an edge-AI platform, and a flagship smartphone. We assess real-time operation through the \emph{duty cycle}, the ratio of inference latency to the prediction stride; since this ratio is also the fraction of the stride budget consumed, a system is real-time feasible when its duty cycle stays below our conservative threshold of $50\%$. With new windows arriving every $1{,}000$\,ms for MHEALTH and $2{,}500$\,ms for USC-HAD, this corresponds to latency budgets of $500$\,ms and $1{,}250$\,ms, respectively.

\solution{} meets this criterion on every hardware target in Table~\ref{tab:deployment}, typically by a wide margin. On the RTX~4090, inference completes in $9.73$\,ms (a duty cycle under $1\%$ on both datasets), and even CPU-only server execution stays comfortably within budget, with the 16- and 4-core x86-64 configurations at $206.4$\,ms and $247.8$\,ms ($\le\!24.8\%$ of the tighter MHEALTH stride). Crucially, feasibility extends well beyond server hardware: the Jetson Xavier NX runs at $35.2$\,ms under TensorRT, and on the Samsung S23 Ultra the ONNX CPU path completes in $377.8$\,ms while the NNAPI GPU path reaches $41.7$\,ms. Taking the S23 Ultra CPU result as a conservative mobile baseline, its duty cycle is $37.78\%$ for MHEALTH and $15.11\%$ for USC-HAD; the GPU path lowers this to $4.17\%$ and $1.67\%$. Full per-device figures are listed in Table~\ref{tab:deployment}.

Overall, \solution{} is not restricted to offline or cloud-hosted inference: it runs in real time on commodity mobile hardware without a server GPU, while leaving substantial headroom for sensor acquisition, preprocessing, OS scheduling, and application logic. This supports the central deployment claim of \solution{}: by translating each sensor window into a small fixed set of latent tokens before LLM inference, the system keeps latency predictable and enables privacy-preserving on-device HAR.



\paragraph{Minimum deployment specification.}
The FP32 checkpoint size of \solution{} is approximately 555\,MB, so a practical deployment should reserve at least 600\,MB of storage for the model weights alone, with additional space for the runtime, tokenizer files, and application code. Based on the validated configurations in Table~\ref{tab:deployment}, the recommended minimum specification for real-time deployment is a modern 4-core CPU or mobile SoC, at least 4\,GB of system memory, and support for an efficient inference runtime such as ONNX Runtime, TensorRT, NNAPI, or an equivalent backend. For best mobile performance, a smartphone-class SoC with an integrated GPU/NPU and 8\,GB or more RAM provides substantial headroom, as demonstrated by the Samsung S23 Ultra results. For embedded edge deployment, an 8\,GB Jetson Xavier NX-class device is sufficient for low-latency accelerated inference. For CPU-only deployment, a 4-core modern processor is sufficient for real-time operation under the evaluated MHEALTH and USC-HAD strides, although accelerator-backed execution is strongly preferred when minimizing energy consumption or supporting additional application workloads. Under these conditions, \solution{} can run locally without cloud inference, preserving sensor-data privacy while meeting real-time HAR constraints.

\paragraph{Future work.}
INT4 LLM-only quantises transformer block weights while keeping the embedding table ($\approx$154\,MB) in FP32. Full INT4 compression of
all layers would reduce the checkpoint to approximately 69\,MB and the total inference footprint to $\sim$144\,MB, enabling deployment on severely RAM-constrained devices such as entry-level wearables. This requires calibrated quantisation (e.g., GPTQ or AWQ) applied to the
embedding table to avoid the NLL-ordering collapse observed with uncalibrated full-model INT8, and is deferred to future work.

The remainder of the on-device feasibility analysis is available in Appendix~\ref{app:ext-on-device-feasibility}.

\subsection{On-device Personalization}
\label{sec:personalization}

\begin{figure}[]
    \centering
    \begin{subfigure}{.49\columnwidth}
        \centering
        \includegraphics[width=\linewidth]{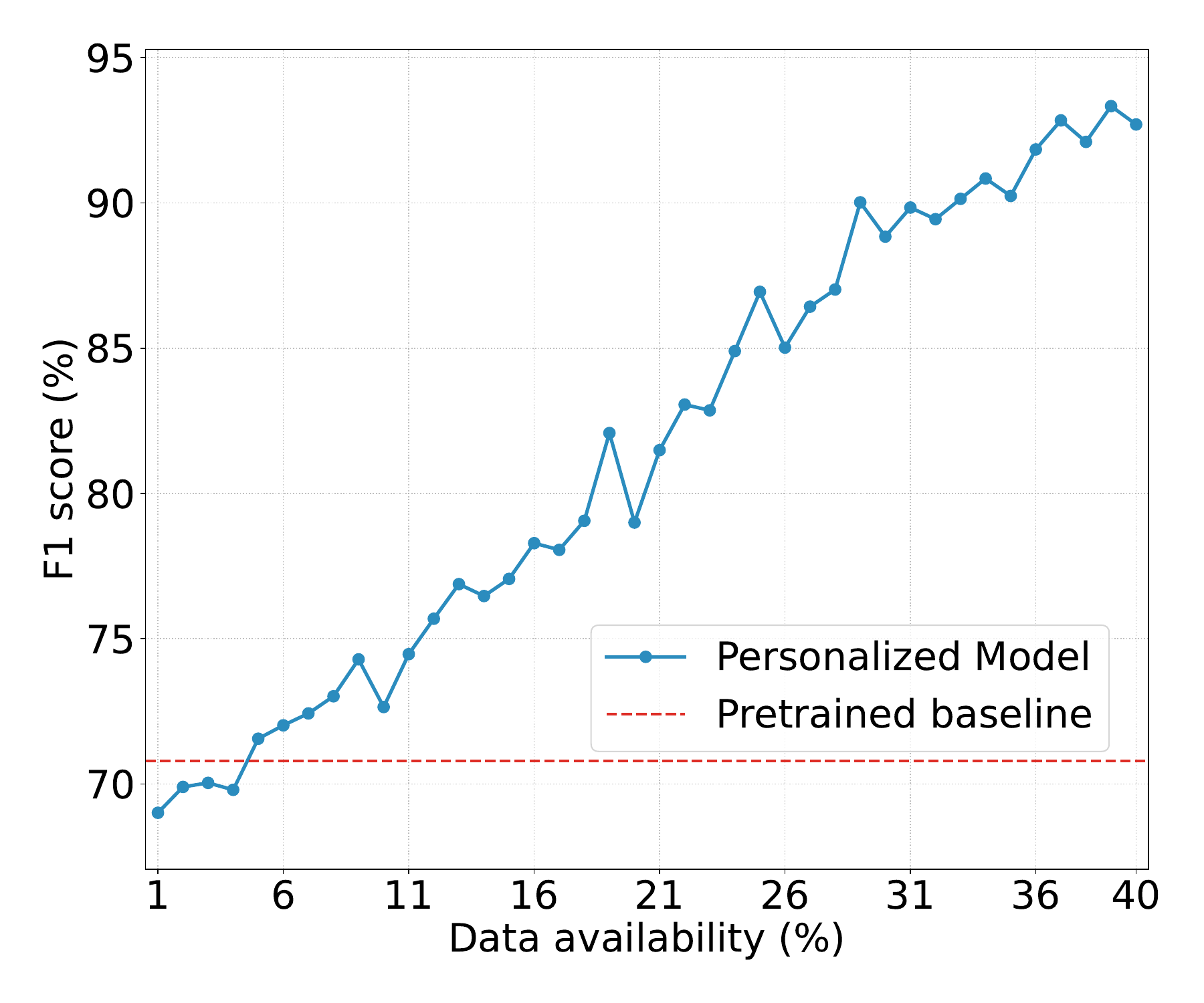}
         \vspace{-5mm}
        \caption{Continued learning for Personalization on USC-HAD}
        \vspace{-2mm}
        \label{fig:personalization_f1}
    \end{subfigure}
    \hfill
    \begin{subfigure}{.49\columnwidth}
        \centering
        \includegraphics[width=\linewidth]{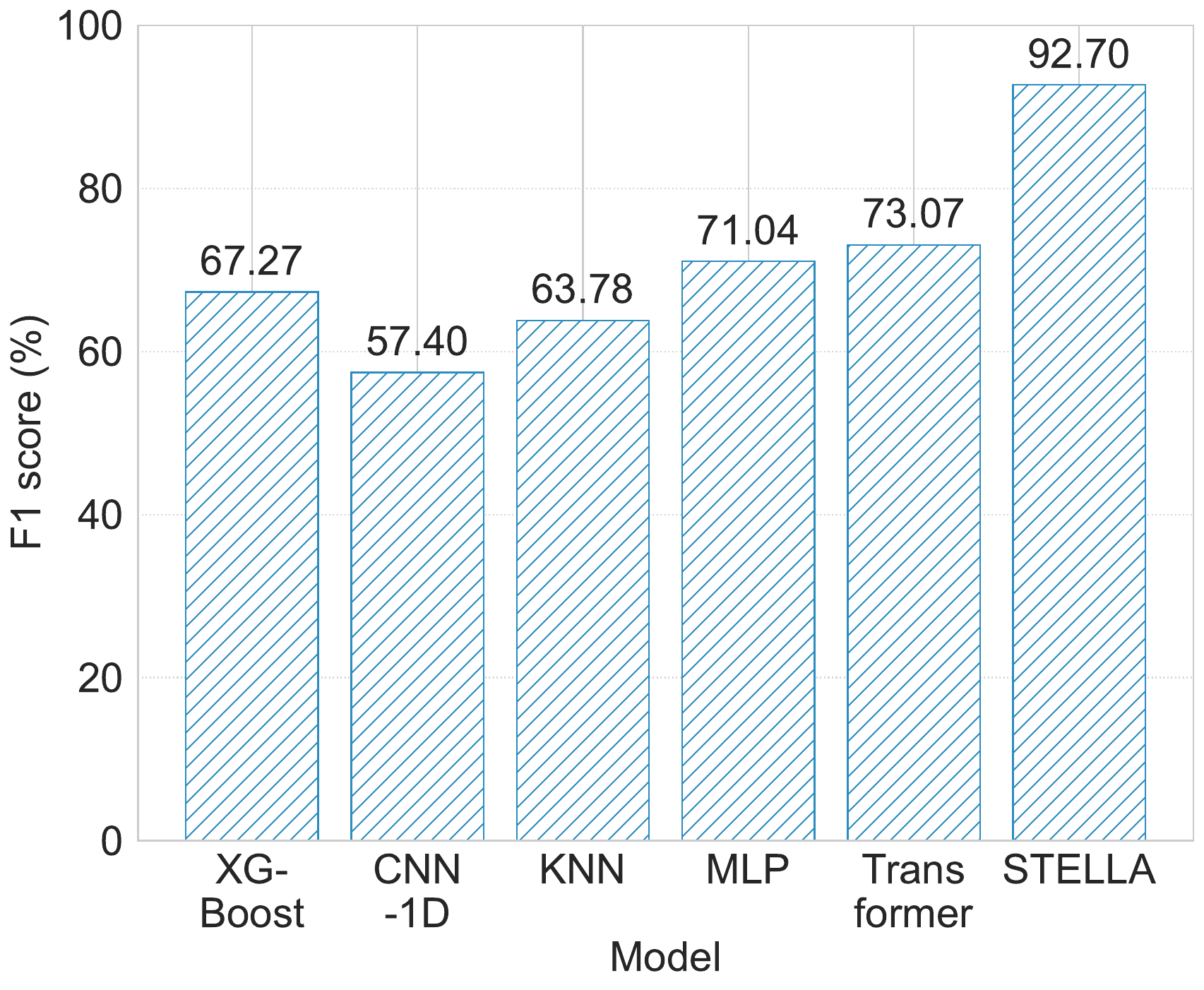}
         \vspace{-5mm}
        \caption{On-device personalization with different models}
        \vspace{-2mm}
        \label{fig:f1-personalized-model}
    \end{subfigure}
\vspace{-2mm}
\caption{On-device Personalization}
\vspace{-6mm}
\label{fig:on-device-person-results}
\end{figure}

\begin{table}[]
\centering
\caption{Personalizing the framework to the user. k=0 case is the baseline. Ret. - Retrieval , RHR - Retrieval Hit Rate. Evaluated on the lowest-performing subject in each dataset to assess whether personalization can improve recognition under the most challenging subject-specific conditions.}
\vspace{-4mm}
\label{tab:personalization}
\small
\setlength{\tabcolsep}{3.5pt}
\begin{tabular}{ccccccccc}
\toprule
& \multicolumn{1}{c}{\textbf{Latency}} &
  \multicolumn{3}{c}{\textbf{MHEALTH}} &
  \multicolumn{3}{c}{\textbf{USC-HAD}} \\
\cmidrule(lr){2-2}\cmidrule(lr){3-5}\cmidrule(lr){6-8}
$k$ &
  \makecell{\textbf{GPU}\\\textbf{ms}} &
  \makecell{\textbf{LLM F1}\\\textbf{(\%)}} &
  \makecell{\textbf{Ret. F1}\\\textbf{(\%)}} &
  \makecell{\textbf{RHR}\\\textbf{(\%)}} &
  \makecell{\textbf{LLM F1}\\\textbf{(\%)}} &
  \makecell{\textbf{Ret. F1}\\\textbf{(\%)}} &
  \makecell{\textbf{RHR}\\\textbf{(\%)}} \\
\midrule
0                   & 9.7   & 95.56   &   -   &   -    & 70.79 &  -    &  -   \\
3                   & 19.0  & 98.90  & 90.74 & 100.00 & 91.10 & 76.96 & 93.84 \\
5                   & 29.1  & 99.80  & 90.74 & 100.00 & 92.08 & 78.33 & 98.63 \\
7                   & 38.2  & 100.00  & 80.13 & 100.00 & 92.49 & 78.70 & 98.63 \\
9                   & 49.0  & 100.00  & 78.02 & 100.00 & 92.70 & 75.83 & 99.32 \\
\bottomrule
\end{tabular}
\vspace{-6mm}
\end{table}

We evaluate whether lightweight on-device personalization can improve recognition for individual users. We hold out the worst-performing subject per dataset (Subject~8 for MHEALTH, Subject~13 for USC-HAD) and split their data into 10\% retrieval context, 40\% fine-tuning, and 50\% test, with non-overlapping windows  to prevent leakage. The LLM remains frozen throughout; only the sensor tokenizer is fine-tuned on the per-user training split, and the retrieval pool is rebuilt from context samples after each epoch. At inference, the top-$k$ context windows (by cosine similarity in sensor-token space) are prepended to the LLM input. We report two auxiliary metrics: \textit{Retrieval F1} (majority vote over the top-$k$ pool, equivalent to a $k$-NN classifier in learned embedding space) and \textit{Retrieval Hit Rate (RHR)} (fraction of test samples whose top-$k$ pool contains at least one correct-class example).

Figure~\ref{fig:personalization_f1} shows the continual personalization trajectory on USC-HAD. Starting from 70.79\% F1 with no user-specific adaptation, performance rises quickly, even 5--10\% of the target user's data already surpasses the pretrained baseline, and peaks at 93.33\% F1 when 39\% of data is available. Table~\ref{tab:personalization} reports final results at the full 40\% personalization split. GPU latency grows from 9.7\,ms ($k=0$) to 49.0\,ms ($k=9$), with retrieval itself contributing less than 1\,ms. The gains are substantial: on MHEALTH, LLM F1 rises from 95.56\% to 100.0\%; on USC-HAD, from 70.79\% to 92.70\% (+21.91 F1 points). On MHEALTH, Retrieval F1 degrades from 90.74\% to 78.02\% as $k$ grows, yet LLM F1 holds at 100.0\% from $k{=}7$, with RHR at 100\% throughout, this confirms the LLM actively reasons over retrieved context rather than performing nearest-neighbour matching. Figure~\ref{fig:f1-personalized-model} compares personalized \solution{} against on-device trainable baselines (KNN, MLP, XGBoost, CNN-1D, Transformer), each given all 50\% of available user labels. \solution{} exceeds all by approximately 20\% F1, confirming that the gains come from the combination of tokenizer adaptation and retrieval-augmented inference, not merely from access to more labelled data.

Overall, these results demonstrate that the proposed framework enables practical on-device personalization through lightweight subject adaptation and retrieval-augmented reasoning. By combining continual accumulation of user-specific knowledge with efficient personalization of the sensor tokenizer, the system progressively adapts to individual users while preserving real-time deployment feasibility and low computational overhead.

\subsection{Ablation Study}
\label{sec:ablation}

We conduct a systematic ablation study across eight design axes of \solution{}, evaluated independently on MHEALTH and USC-HAD under identical
training conditions (40~epochs, lr\,$= 5\!\times\!10^{-4}$, batch\,$= 16$, $N{=}16$ tokens, seed\,$= 42$). Consistent reporting on two datasets with contrasting sensor configurations (21-channel multi-placement vs.\ 6-channel single-IMU) reveals which design choices are universally beneficial and which are dataset-dependent. Table~\ref{tab:ablation} summarises all results; default configurations are shown in bold. We perform a 5-fold held-out evaluation protocol for these experiments.

\begin{table}[t]
\centering
\caption{Ablation study across all design axes (5-fold LOSO). Bold rows denote the default configuration for each group. The remainder of the ablation study is available in Appendix~\ref{app:ext-ablation}.}
\vspace{-4mm}
\label{tab:ablation}
\footnotesize
\setlength{\tabcolsep}{1.5pt}
\begin{tabular}{cll rr rr r}
\toprule
\textbf{Grp} & \textbf{ID} & \textbf{Variant} &
  \multicolumn{2}{c}{\textbf{MHEALTH}} &
  \multicolumn{2}{c}{\textbf{USC-HAD}} &
  \textbf{Trainable} \\
\cmidrule(lr){4-5}\cmidrule(lr){6-7}
 & & &
  \makecell{\textbf{Acc.}\\\textbf{(\%)}} &
  \makecell{\textbf{F1}\\\textbf{(\%)}} &
  \makecell{\textbf{Acc.}\\\textbf{(\%)}} &
  \makecell{\textbf{F1}\\\textbf{(\%)}} &
  \textbf{Params} \\
\midrule
\multirow{4}{*}{A}
  & A1 & Attention output only (H$_4$)               & 96.27 & 95.79 & 86.08 & 79.34 & 14.11\,M \\
  & A2 & Stage-3 only (H$_3$)                        & 96.22 & 95.57 & 86.40 & 80.17 & 14.11\,M \\
  & A3 & Stage-3 $+$ attention (H$_3$+H$_4$)        & 96.29 & 95.57 & 86.23 & 79.41 & 14.26\,M \\
  & \textbf{A4} & \textbf{Full fusion H$_2$+H$_3$+H$_4$} & \textbf{96.68} & \textbf{96.22} & \textbf{86.92} & \textbf{81.01} & \textbf{14.34\,M} \\
\midrule
\multirow{2}{*}{B}
  & B1 & Full attention in encoder stages       & 96.94 & 96.54 & 86.23 & 79.29 & 18.42\,M \\
  & \textbf{B2} & \textbf{RepMixer stages} & \textbf{96.68} & \textbf{96.22} & \textbf{86.92} & \textbf{81.01} & \textbf{14.34\,M} \\
\midrule
\multirow{3}{*}{C}
  & C1 & Shallow depths $(1,2,3)$                    & 96.72 & 96.19 & \textbf{87.37} & \textbf{82.09} & 10.07\,M \\
  & \textbf{C2} & \textbf{Default depths $(2,4,6)$}  &
    \textbf{96.68} & \textbf{96.22} & 86.92 & 81.01 & \textbf{14.34\,M} \\
  & C3 & Deep depths $(3,6,9)$                       & 96.51 & 95.91 & 86.04 & 80.26 & 18.60\,M \\
\bottomrule
\end{tabular}
\vspace{-4mm}
\end{table}

\subsubsection{\textbf{Multi-Scale Token Aggregation (Group~A)}}
\label{sec:abl-fusion}

Group~A evaluates the contribution of each hierarchical representation. Full multi-scale fusion (A4: H$_2$+H$_3$+H$_4$) provides the strongest overall configuration, and removing any scale consistently reduces F1 across both datasets. Compared with the next-best fused variant, A3 (H$_3$+H$_4$), full fusion improves F1 by 0.65 points on MHEALTH and 1.60 points on USC-HAD. The larger gain on USC-HAD suggests that the coarse-grained H$_2$ representation is especially useful for sparse single-IMU data, where global temporal structure carries more discriminative information. This confirms that multi-scale fusion is a universally beneficial design choice for \solution{}.

\subsubsection{\textbf{RepMixer vs. Self-Attention (Group~B)}}
\label{sec:abl-mixer}
The cross-dataset comparison provides a definitive resolution of the RepMixer vs.\ self-attention trade-off. On MHEALTH, B1 (full attention)
marginally outperforms RepMixer (B2) by 0.32\,F1 points. On USC-HAD, the relationship \emph{inverts}: B2 (RepMixer) outperforms B1 by \textit{1.72\,F1 points}. Taken together across both datasets, RepMixer is the consistently stronger or equally competitive choice, while using 22.2\% fewer tower parameters (14.34\,M vs.\ 18.42\,M). The inversion on USC-HAD is attributed to the longer 500-sample input sequences, where early-stage self-attention incurs higher quadratic cost and has less benefit over the locally periodic convolutional inductive bias suited to single-IMU signals. We retain RepMixer (B2) as the universal default.

\subsubsection{\textbf{Encoder Depth (Group~C)}}
\label{sec:abl-depth}

The depth ablation produces a consistent result: adding parameters beyond the default depth does not help, and the shallow encoder is competitive or superior. On MHEALTH, C1 (shallow) and C2 (default) are statistically indistinguishable (96.72\% vs.\ 96.68\%, within $\pm$4.01\% cross-fold
standard deviation). On USC-HAD, C1 outperforms C2 by a larger margin (87.37\% vs.\ 86.92\% accuracy; 82.09\% vs.\ 81.01\% F1), suggesting that
for single-IMU signals with higher activity ambiguity, the stronger regularisation provided by a shallower encoder is beneficial. The deep
encoder C3 underperforms both across datasets. The practical implication is clear: the shallow configuration (C1, 10.07\,M trainable parameters) delivers equal or better accuracy than the default (C2, 14.34\,M) at a 30\% parameter reduction, and should be preferred for memory-constrained deployments.

\subsubsection{Summary}
\label{sec:abl-summary}


The cross-dataset ablation yields five universally stable design principles and two dataset-sensitive findings. \textit{Universally stable}: (i)~full multi-scale fusion (A4) is consistently the strongest aggregation strategy, confirming the value of combining H$_2$, H$_3$, and H$_4$ representations; (ii)~RepMixer (B2) provides the most robust encoder trade-off, matching or approaching self-attention on MHEALTH while outperforming it on the harder USC-HAD benchmark with fewer parameters; (iii)~increasing encoder depth beyond the default brings no accuracy benefit and often hurts, while the shallow encoder remains competitive and may be preferable for memory-constrained deployment; (iv)~freezing the LLM (D1) is more reliable than full or partial fine-tuning, especially on the sparse-sensor USC-HAD dataset; and (v)~the LLM-based NLL scoring head is essential, as replacing it with a conventional linear classifier causes substantial drops on both datasets.

\textit{Dataset-sensitive}: semantic label benefit depends on sensor richness, improving MHEALTH more clearly than USC-HAD; and LLM backbone scaling improves performance most reliably within the GPT family, while non-GPT backbones do not provide consistent gains despite similar or larger parameter counts. Together, these findings show that \solution{} is strongest when the sensor tokenizer, multi-scale fusion, frozen GPT-style language backbone, task prompting and NLL-based activity scoring are retained as a unified design. GPT-2 remains the best efficiency--performance default, while larger GPT variants can be used when additional compute and memory are available.

%% file: Files/main-table.tex
\begin{table}[] 
\centering
\scriptsize
\caption{Benchmarking  \solution{} with SOTAs. The reference shows from where the values were taken.}
\vspace{-4mm}
\label{tab:results-comparison-table}
\renewcommand{\arraystretch}{1.2}
\begin{tabular}{|l|l|l|l|}
\hline
\textit{\textbf{Dataset}} & \textit{\textbf{Model}}                         & \textit{\textbf{Acc(\%)}} & \textit{\textbf{F1}(\%)}   \\ \hline
\multirow{6}{*}{USC-HAD}   & Triplet LSTM baseline ~\cite{khaertdinov2021deep}    & -                          & 53.50      \\ \cline{2-4} 
                          & DeepConvLSTM   ~\cite{khaertdinov2021deep}      & -                          & 46.00  \\ \cline{2-4} 
                          & Transformer-like architecture ~\cite{khaertdinov2021deep}         & -            & 55.00    \\ \cline{2-4}      
                          & Triplet LSTM (HTL-SB) ~\cite{khaertdinov2021deep} [Percom '21] & -     & 62.80    \\ \cline{2-4} 
                          & HARGPT~\cite{ji2024hargpt} & 28.90  & 21.00 \\ \cline{2-4}
                          & LLM as VA~\cite{hota2025evaluating} & 34.51 & 32.10 \\ \cline{2-4}
                          & SensorLLM ~\cite{li2025sensorllm} [EMNLP '25]  & 62.60 & 61.20 \\ \cline{2-4}
                          & RAG-HAR~\cite{sivaroopan2025rag} [Percom '26]                  & 57.20                & 58.63   \\ \cline{2-4}
                          & \textbf{\solution{}  (ours)}                  & \textbf{76.36}             & \textbf{72.00}   \\ \hline
\multirow{6}{*}{HHAR}       & CPC ~\cite{haresamudram2023investigating}   & -                          & 59.17 \\ \cline{2-4} 
                          & GRU Classifier ~\cite{haresamudram2023investigating}        & -                          & 45.23     \\ \cline{2-4} 
                          & DeepConvLSTM ~\cite{haresamudram2023investigating}        & -                          & 52.37    \\ \cline{2-4} 
                          & SimCLR ~\cite{haresamudram2023investigating}   & -                & 52.84            \\ \cline{2-4}      
                          & Enhanced CPC ~\cite{haresamudram2023investigating}    [Percom '23]     & -     & 59.25   \\ \cline{2-4} 
                          & RAG-HAR~\cite{sivaroopan2025rag} [Percom '26]       & 58.61                & 59.86      \\ \cline{2-4}
                          & \textbf{\solution{}  (ours)}       & \textbf{72.71}                & \textbf{71.69}      \\ \hline
\multirow{8}{*}{MHEALTH} & MC-CNN  ~\cite{suh2022adversarial}           & 89.69               & 87.42  \\ \cline{2-4} 
                          & DeepConvLSTM ~\cite{suh2022adversarial}   & 89.24              & 87.17 \\ \cline{2-4} 
                          & Transformer-like ~\cite{suh2022adversarial}    & 87.44    & 85.13  \\ \cline{2-4} 
                          & METIER  ~\cite{suh2022adversarial}      & 94.42              & 94.09  \\ \cline{2-4} 
                          & ADFE ~\cite{suh2022adversarial}    [Percom '22]        & 96.72         & 96.47  \\ \cline{2-4} 
                          & Triplet LSTM (HTL-SB) ~\cite{khaertdinov2021deep} [Percom '21]  & -                & 65.60              \\ \cline{2-4}
                          & HARGPT~\cite{ji2024hargpt} & 24.00  & 22.05 \\ \cline{2-4}
                          & LLM as VA~\cite{hota2025evaluating} & 32.78 & 30.11 \\ \cline{2-4}
                          & SensorLLM ~\cite{li2025sensorllm} [EMNLP '25] & 89.00 & 89.40 \\ \cline{2-4}
                          & RAG-HAR~\cite{sivaroopan2025rag} [Percom '26]      & 96.91              & 96.74         \\ \cline{2-4}
                          & \textbf{\solution{}  (ours)}         & \textbf{97.93}              & \textbf{97.15}         \\ \hline
\multirow{7}{*}{GOTOV}    & LSTM Learner Baseline  ~\cite{ahmad2023alae}         & -                          & 61.10            \\ \cline{2-4} 
                          & DeepConvLSTM ~\cite{ahmad2023alae}                   & -                          & 66.90              \\ \cline{2-4} 
                          & b-LSTM-S ~\cite{ahmad2023alae}                & -                          & 63.90         \\ \cline{2-4} 
                          & Att. Model ~\cite{ahmad2023alae}             & -                          & 70.70       \\ \cline{2-4} 
                          & Attend and Discriminate ~\cite{ahmad2023alae}     & -                          & 76.20             \\ \cline{2-4} 
                          & ALAE-TAE-CutMix+  ~\cite{ahmad2023alae} [Percom '23]      & -           & \underline{79.40}  \\ \cline{2-4} 
                          & RAG-HAR~\cite{sivaroopan2025rag} [Percom '26]      & 81.55        & 79.92                         \\ \cline{2-4}
                          & \textbf{\solution{} (ours)}      & \textbf{83.15}        & \textbf{82.60}                         \\ \hline
\multirow{7}{*}{SKODA}    & LSTM Learner Baseline ~\cite{ahmad2023alae}  & -                          & 90.40        \\ \cline{2-4} 
                          & DeepConvLSTM ~\cite{ahmad2023alae}   & -                          & 91.20      \\ \cline{2-4} 
                          & b-LSTM-S  ~\cite{ahmad2023alae}      & -                          & 92.10             \\ \cline{2-4} 
                          & Att. Model~\cite{ahmad2023alae}         & -                          & 91.30            \\ \cline{2-4} 
                          & Attend and Discriminate  ~\cite{ahmad2023alae}       & -                          & 92.80              \\ \cline{2-4} 
                          & ALAE-TAE-CutMix+  ~\cite{ahmad2023alae} [Percom '23]   & -             & 94.80   \\ \cline{2-4} 
                          & RAG-HAR~\cite{sivaroopan2025rag} [Percom '26]     & 96.04                & 95.21       \\ \cline{2-4}
                          & \textbf{\solution{} (ours)}     & \textbf{97.37}                & \textbf{95.87}       \\ \hline
\multirow{7}{*}{Opportunity (locomotion)}    & DWT ~\cite{li2021units}   & 87.43               & 89.01    \\ \cline{2-4} 
                          & ResNet ~\cite{li2021units}   & 89.54                         & 91.36      \\ \cline{2-4} 
                          & MaCNN ~\cite{li2021units}      & 88.32                          & 90.32             \\ \cline{2-4} 
                          & SenseHAR ~\cite{jeyakumar2019sensehar} [SenSys '19]       & 88.90      & 88.60            \\ \cline{2-4} 
                          & THAT ~\cite{li2021units}   & 90.35         & 92.12   \\ \cline{2-4} 
                          & LaxCat ~\cite{li2021units}    & 60.19           & 41.47   \\ \cline{2-4} 
                          & Uni-TS ~\cite{li2021units} [SenSys '21]     & 91.58                & 93.33       \\ \cline{2-4}
                          & \textbf{\solution{} (ours)}     & \textbf{97.27}                & \textbf{94.29}       \\ \hline
\multirow{7}{*}{Opportunity (gesture)}    & DNN ~\cite{jeyakumar2019sensehar}   & 67.21     & 58.42    \\ \cline{2-4} 
                          & CNN ~\cite{jeyakumar2019sensehar}   & 71.13                         & 64.31      \\ \cline{2-4} 
                          & Conv-LSTM  ~\cite{jeyakumar2019sensehar}     & 74.80      & 67.92             \\ \cline{2-4} 
                          & MA-DNN ~\cite{jeyakumar2019sensehar}      & 72.86      & 66.84            \\ \cline{2-4} 
                          & MA-CNN ~\cite{jeyakumar2019sensehar}       & 74.40                  & 67.54              \\ \cline{2-4} 
                          & SenseHAR ~\cite{jeyakumar2019sensehar} [SenSys '19]     & 74.22                & 67.48       \\ \cline{2-4}
                          & \textbf{\solution{} (ours)}     & \textbf{79.88}                & \textbf{69.75}       \\ \hline
\multirow{7}{*}{PAMAP2}    & DNN ~\cite{jeyakumar2019sensehar}  & 85.82     & 86.22    \\ \cline{2-4} 
                          & CNN  ~\cite{jeyakumar2019sensehar}  & 90.40     & 90.36      \\ \cline{2-4} 
                          & Conv-LSTM  ~\cite{jeyakumar2019sensehar}     & 94.68      & 94.46             \\ \cline{2-4} 
                          & MA-DNN ~\cite{jeyakumar2019sensehar}      & 92.16      & 92.04            \\ \cline{2-4} 
                          & MA-CNN~\cite{jeyakumar2019sensehar}        & 95.14                  & 94.99              \\ \cline{2-4} 
                          & SenseHAR ~\cite{jeyakumar2019sensehar} [SenSys '19]     & 95.32                & 95.08       \\ \cline{2-4}
                          & \textbf{\solution{} (ours)}     & \textbf{96.48}                & \textbf{96.52}       \\ \hline

\end{tabular}
\vspace{-7mm}
\end{table}

%% file: Files/Discussion.tex
\section{Discussion}

\subsection{Local LLM at the Edge}

\solution{} is motivated by the trend toward small local LLMs that serve as shared, system-level backends on mobile and edge devices for text, voice, and multimodal reasoning. By confining all HAR-specific learning to the tokenizer, \solution{} lets sensing reuse this backend rather than adding a separate model: the sensor tokenizer plays the role for sensor windows that the LLM's native tokenizer plays for text, making sensing one more modality the frozen LLM reasons over. A device already hosting a local LLM thus needs only to add this compact front-end, instead of fine-tuning or rebuilding the LLM per task. This reflects a broader direction for edge intelligence, attach modality-specific tokenizers to a common LLM and reuse its semantic reasoning across applications.

\subsection{NLL Scoring vs. Free-form Generation}

\solution{} recognizes activities by scoring candidate labels under the LLM's likelihood (selecting the lowest-NLL label) rather than generating a label as free text, which matters for on-device use in three ways. First, generation would require a longer, more instructive prompt (enumerating candidate activities and formatting rules) whose extra tokens inflate attention and KV-cache cost, working against our goal of a compact input. Second, small on-device LLMs generate unreliably: they may emit synonyms, explanations, or labels outside the dataset, requiring post-processing and adding avoidable errors, whereas scoring restricts predictions to valid labels by construction. Third, scoring cost is predictable: it grows only with the (typically small) number of candidate labels, the sensor-token prefix is shared across candidates and can be batched, and personalization can further restrict scoring to the labels present in the retrieved context. NLL scoring therefore keeps the prompt short, avoids invalid outputs, and yields predictable latency.

\subsection{Energy Optimization for On-device Use}

Because \solution{}'s per-inference cost is set by the token budget rather than the raw signal, it exposes a few knobs for matching a device's real-time and energy budget without touching the LLM. Sustained energy is governed mainly by the prediction stride, since continuous-HAR energy scales with the duty cycle; lengthening the stride lowers average power at the cost of temporal resolution. Per-inference cost is set by the token budget $N$ and execution precision: smaller $N$ and FP16/INT4 quantization both reduce latency and memory with only graceful accuracy loss. Window length and sampling rate, by contrast, trade accuracy against sensor-acquisition energy but barely affect LLM latency, since the model always sees $N$ tokens, so low-channel deployments can use longer windows or lower sampling rates to save energy while even improving accuracy. Finally, accelerator-backed execution cuts active time by roughly an order of magnitude over CPU, sharply lowering the duty cycle. A deployment therefore selects the operating point ($N$, window, stride, precision, backend) that fits its constraints, and the fixed-token interface keeps this recipe largely independent of channel count and window length. These trade-offs are quantified in our on-device feasibility analysis (Section~\ref{sec:feasibility}).

%% file: Files/Conclusion.tex
\section{Conclusion}

This paper presented \solution{}, an efficient sensor-to-LLM translation framework for on-device human activity recognition. Rather than treating LLM-based HAR as a prompting problem or a model-compression problem, \solution{} reframes it as a sensor tokenization problem: sensor time-series data should be translated into compact, activity-aware tokens before reaching the LLM. Through a lightweight hierarchical tokenizer, frozen LLM integration, and local personalization mechanism, \solution{} shows that language models can be used for HAR without sacrificing edge efficiency, privacy, or reusability. The evaluation demonstrates that this design consistently improves recognition performance, produces stronger sensor representations than generic time-series tokenizers, and remains practical for real-time on-device deployment. Overall, \solution{} provides a promising step toward accurate, private, and personalized LLM-based sensing systems on future mobile and edge platforms.

%% file: Files/Appendix.tex
\section{Datasets}
\label{app:dataset}

We use publicly available and widely used HAR datasets for our experiments. Since prior HAR benchmarks commonly use dataset-specific evaluation protocols, we follow the established preprocessing and split settings used by the corresponding prior work for each dataset. Accordingly, results should be interpreted as within-dataset comparisons under the same protocol used by the cited baselines, rather than as a single unified cross-dataset split policy. For all datasets, train/validation/test separation is performed before model fitting and evaluation. Normalization statistics are computed only from the training data and then applied to validation and test data. Validation and test samples are not used for training, hyperparameter selection, retrieval-memory construction, or personalization. For datasets using sliding windows, we first apply the dataset split at the recording, subject, trial, or temporal-segment level specified by the protocol, and then generate windows separately within each split. This prevents overlapping windows from crossing train/validation/test boundaries. To make the split unit explicit, HHAR and MHEALTH use subject-independent LOSO evaluation; PAMAP2 and USC-HAD use subject-independent held-out-subject evaluation; GOTOV uses subject-independent held-out-user validation and test splits; Opportunity uses trial- or recording-file-level splits depending on the task; and SKODA uses a chronological recording-level split. None of our experiments use random window-level splitting across the full set of overlapping windows. For all datasets with overlapping windows, the split is applied before sliding-window segmentation, and windows are generated independently within each split. Therefore, overlapping windows do not cross train/validation/test boundaries.

\subsection{HHAR~\cite{stisen2015smart}}
This dataset primarily contains locomotion-style activities such as walking, sitting, standing, and going up/down stairs. The dataset consists of sensor data collected from smartphones and smartwatches. Following~\cite{haresamudram2023investigating}, we use only wrist-worn sensor data, which consists of 6 channels. The window size is set to 2 seconds with 50\% overlap. The evaluation protocol is leave-one-subject-out (LOSO). Data preparation and splits follow~\cite{haresamudram2023investigating}. Since the protocol is subject-disjoint, data from the held-out subject is not used during training or validation.

\subsection{PAMAP2~\cite{reiss2012introducing}}
The dataset originally contains 52 channels, including heart rate and full IMU sensor streams. Following~\cite{jeyakumar2019sensehar}, we retain the three IMUs located on the wrist, ankle, and chest. For each IMU, we use accelerometer, gyroscope, and magnetometer streams, resulting in 27 channels in total. We focus on 12 protocol-defined activities recorded from 9 subjects. In addition to locomotion activities, PAMAP2 contains daily living activities such as ironing and vacuum cleaning. The IMU data was sampled at 100 Hz, and we apply a sliding window of 500 samples (5 seconds) with a step size of 100 samples (1 second). Following the split in~\cite{jeyakumar2019sensehar}, subject 4 is used as the test set and the remaining subjects are used for training. Thus, the test set is subject-disjoint from the training data.

\subsection{MHEALTH~\cite{banos2014mhealthdroid}}
The dataset was sampled at 50 Hz and contains 21 channels. In addition to locomotion activities, it includes exercise activities such as jogging and cycling. The dataset contains 12 activities in total. Following~\cite{suh2022adversarial}, we use sliding windows of 200 samples (4 seconds) with a step size of 50 samples (1 second). The evaluation protocol is leave-one-subject-out (LOSO). Data preparation and splits follow~\cite{suh2022adversarial}. In each LOSO fold, data from the held-out subject is used only for testing, while the remaining subjects are used for training and validation.

\subsection{GOTOV~\cite{paraschiakos2020activity}}
The dataset contains 9 sensor channels collected from three body positions on 35 older adults aged over 61 while performing 16 activities. This dataset contains the same activities performed at different paces, such as slow walking, normal walking, and fast walking, as well as the same activity performed with different equipment, such as sitting on a sofa, sitting on a chair, and sitting on a couch. Six participants are excluded due to missing sensor data. Following~\cite{ahmad2023alae}, data from 3 users is used for testing and data from 3 users is used for validation, each consisting of two males and one female, while data from the remaining users is used for training. Windows are then created within each split using 24-sample segments with 50\% overlap between consecutive windows. This protocol keeps training, validation, and test users disjoint. Data preparation and splits follow~\cite{ahmad2023alae}.

\subsection{SKODA~\cite{stiefmeier2008wearable}}
This dataset contains recordings of 10 assembly-line activities performed by car manufacturing workers, captured through 60 sensor channels positioned on the right-hand side of the body. Following~\cite{ahmad2023alae}, the recording is split chronologically, with the first 80\% used for training, the next 10\% used for validation, and the remaining 10\% used for testing. After this split is applied to the recording, windows are created separately within each split using 24-sample segments with 50\% overlap between consecutive windows. This ensures that overlapping windows do not cross train/validation/test boundaries. Data preparation and splits follow~\cite{ahmad2023alae}.

\subsection{USC-HAD~\cite{zhang2012usc}}
This dataset includes accelerometer and gyroscope signals, each with three degrees of freedom, resulting in six channels per time step. Data was collected from 14 subjects performing 12 basic activities such as walking in different directions, running, and jumping. Following~\cite{khaertdinov2021deep}, the dataset is segmented into 1-second windows with 50\% overlap. Data from subjects 13 and 14 is reserved for testing, while the remaining subjects are used for training. Therefore, the test set is subject-disjoint from the training data. Data preparation and splits follow~\cite{khaertdinov2021deep}.

\subsection{Opportunity~\cite{roggen2010collecting}}
The dataset consists of four trials and includes measurements from a wide range of sensing modalities, such as body-worn sensors, object-mounted sensors, and ambient sensors. Following~\cite{jeyakumar2019sensehar, li2021units}, we use only the five body-worn inertial measurement units placed at different body locations: the left lower arm (LLA), left upper arm (LUA), right lower arm (RLA), right upper arm (RUA), and back torso. Each inertial unit records accelerometer, gyroscope, and magnetometer signals. The dataset provides activity annotations at two levels of granularity. The high-level labels correspond to major locomotion states, including sitting, standing, walking, lying, and random movement. The low-level gesture labels describe 17 fine-grained manipulation activities, such as opening and closing doors, shelves, and drawers, as well as drinking tea. For gesture evaluation, we follow the data preparation protocol in~\cite{jeyakumar2019sensehar}. We discard samples with null activity labels, use Trials 1--3 for training, and reserve Trial 4 for testing. The dataset is resampled to 30 Hz and segmented using windows of 256 timesteps. For locomotion evaluation, we follow the data preparation protocol of~\cite{li2021units}. Specifically, we use 20 recording files to construct the training split and the remaining four files for the test split. We retain the original sampling rate and segment the sensor streams using 5-second windows with a 1-second overlap. For both gesture and locomotion evaluation, windowing is performed after the trial or recording-file split is applied, so windows do not cross split boundaries.

\section{Tokenizer Baseline Implementation Details}
\label{app:tokenizer_details}

This section provides implementation details for the tokenizer comparison in Section~\ref{sec:tokenizer}. The purpose of this experiment is to isolate the effect of the sensor tokenizer rather than changes in the downstream LLM, token budget, or training procedure. Therefore, all tokenizer variants are evaluated inside the same \solution{} pipeline. Each encoder receives the same preprocessed sensor windows, outputs the same number of latent sensor tokens, and is followed by the same two-layer MLP projector into the LLM embedding space. The downstream LLM is GPT-2 and remains frozen for all variants. Only the sensor tokenizer and projector are trained. All tokenizers are constrained to output $N=16$ sensor tokens. We use the same language-model classification objective, optimizer, learning rate, batch size, dropout, number of epochs, and evaluation protocol across all variants. Unless otherwise stated, models are trained for 40 epochs with AdamW, learning rate $5 \times 10^{-4}$, weight decay $10^{-4}$, batch size 16, dropout 0.1, two warm-up epochs, cosine learning-rate decay, and early stopping with patience 6. The reported results use the same held-out evaluation protocol as the tokenizer comparison in Section~\ref{sec:tokenizer}.

Table~\ref{tab:tokenizer_impl_details} summarizes the encoder variants used in the tokenizer comparison. The parameter count refers only to the sensor tokenizer unless otherwise specified. The projector and frozen GPT-2 backbone are identical across variants and are therefore not used to distinguish the tokenizer baselines.

\begin{table}[t]
\centering
\small
\caption{Implementation details for tokenizer baselines. All variants are reimplemented and evaluated with the same frozen GPT-2 backend, the same two-layer MLP projector, and the same output token budget of $N=16$.}
\vspace{-4mm}
\label{tab:tokenizer_impl_details}
\begin{tabularx}{\linewidth}{l X}
\toprule
Variant & Encoder configuration \\
\midrule
STELLA-Tower & Multi-scale RepMixer tower, dims $(96,192,384)$, depths $(2,4,6)$ \\
PatchTST & Patch embedding + Transformer, $d=384$, 8 heads, 3 layers, patch length 16 \\
Chronos-style & Mean-scale normalization + patch Transformer, $d=384$, 8 heads, 3 layers, patch length 16 \\
TimesNet & FFT-based period detection with 3 periods and 2 Inception-style layers \\
iTransformer & Inverted channel-as-token Transformer, $d=384$, 8 heads, 3 layers, temporal pool size 64 \\
TSMixer & Alternating temporal and channel MLP mixing, $d=384$, 4 layers, temporal pool size 64 \\
TimeMixer & Multi-scale decomposable MLP mixing, $d=384$, 4 scales, 3 layers \\
\bottomrule
\end{tabularx}
\vspace{-6mm}
\end{table}

For Chronos, we do not compare against different released Chronos checkpoint sizes. Instead, we use a Chronos-style patch Transformer encoder with the same hidden dimension, token budget, and downstream LLM pipeline as the other tokenizer baselines. This avoids confounding the tokenizer comparison with differences in pretrained checkpoint size or external pretraining data. Thus, the comparison should be interpreted as a controlled comparison of tokenizer architectures under a matched \solution{}-style LLM classification pipeline, not as a comparison against every publicly released Chronos model variant. Since GPT-2 is frozen and shared across all variants, differences in accuracy and encoder latency mainly reflect the sensor-tokenization architecture.

\section{Prompts}
\label{app:prompts}

For reproducibility, this appendix documents the exact prompts  used by STELLA across all datasets and inference settings. \solution{} uses two prompt templates: one for the standard (non-personalized) inference setting and one for the personalized setting where retrieved context examples are prepended to the LLM input.

\textbf{Standard inference prompt.} In the standard setting, the sensor tokens are followed by a short instruction prompt:
\begin{quote}
\texttt{Classify the activity from the sensor window.}\\
\texttt{Answer:}
\end{quote}

\textbf{Personalized inference prompt.} In the on-device personalization setting (Section~\ref{sec:personalization}), the top-$k$ retrieved context examples are inserted before the query sensor tokens, and the prompt is modified to reference them:
\begin{quote}
\texttt{The following are labeled sensor examples. Using them as reference, classify the current sensor window.}\\
\texttt{Answer:}
\end{quote}

The same template is used across all seven datasets without dataset-specific tuning, isolating performance differences from prompt engineering.

\section{Extended On-device feasibility analysis}
\label{app:ext-on-device-feasibility}

\subsection{\textbf{Effect of Input Window Duration}}
\label{sec:window}

Table~\ref{tab:window} evaluates classification performance and inference latency across a range of input window durations on both datasets. For
MHEALTH (50\,Hz), windows span 1--6\,s; for USC-HAD (100\,Hz), they span 1--10\,s.

\begin{table}[t]
\centering
\caption{Accuracy and latency vs.\ window duration. Default configurations shown in bold.}
\vspace{-4mm}
\label{tab:window}
\small
\setlength{\tabcolsep}{3.5pt}
\begin{tabular}{llrrrrrr}
\toprule
\textbf{DS} & \textbf{Window} & \textbf{Samples} &
  \makecell{\textbf{Acc.}\\\textbf{(\%)}} &
  \makecell{\textbf{F1}\\\textbf{(\%)}} &
  \makecell{\textbf{GPU}\\\textbf{p50 (ms)}} &
  \makecell{\textbf{CPU 4T}\\\textbf{p50 (ms)}} \\
\midrule
MH & 1\,s  &  50  & 94.75 & 94.16 &  7.72 & 221.2 \\
MH & 2\,s  & 100  & 96.38 & 95.88 & 10.24 & 226.7 \\
MH & \textbf{4\,s} & \textbf{200} & \textbf{96.68} & \textbf{96.22} & \textbf{10.31} & \textbf{229.7} \\
MH & 6\,s  & 300  & 95.85 & 95.35 & 10.54 & 234.4 \\
\midrule
USC & 1\,s  & 100  & 81.68 & 77.29 &  7.86 & 197.9 \\
USC & 2\,s  & 200  & 84.56 & 79.99 &  8.49 & 226.0 \\
USC & \textbf{5\,s} & \textbf{500} & \textbf{86.92} & \textbf{81.01} & \textbf{8.95} & \textbf{205.4} \\
USC & 10\,s & 1000 & 89.74 & 83.40 & 16.00 & 226.7 \\
\bottomrule
\end{tabular}
\vspace{-6mm}
\end{table}

Two findings are of note. First, the CPU latency remains nearly constant across all window durations on both datasets: MHEALTH spans only 13.5\,ms
across a 6$\times$ input size range (221.2\,$\to$\,234.4\,ms), and USC-HAD spans 28.8\,ms across a 10$\times$ range (197.9\,$\to$\,226.7\,ms). This
confirms that temporal compression to a fixed $N{=}16$ token sequence renders LLM inference --- the dominant cost --- invariant to window length.

Second, the optimal window duration is dataset-dependent and inversely related to sensor richness. MHEALTH peaks at 4\,s and degrades beyond:
21-channel multi-placement sensors provide sufficient spatial redundancy that longer windows introduce more label-mixing than useful context.
USC-HAD, equipped with only 6 channels from a single IMU, continues to improve up to 10\,s: the model compensates for reduced modality richness by
integrating longer temporal context to resolve activity ambiguities (e.g., distinguishing walking left from walking right requires observing more
stride cycles). The practical implication is that sparse-sensor deployments should favour longer acquisition windows, and \solution{} accommodates this at negligible CPU latency cost.

\subsection{\textbf{Robustness to Reduced Sampling Rate}}
\label{sec:samplerate}

Table~\ref{tab:samplerate} evaluates both datasets under two sampling-rate reduction strategies with Butterworth anti-aliasing (order~4).
\textit{Approach~A} (dataset-level decimation) preserves the fixed number of model-input samples while extending the real-time context covered.
\textit{Approach~B} (window-level decimation) preserves the real-time context duration while reducing temporal resolution.

\begin{table}[t]
\centering
\caption{Accuracy under reduced sensor sampling rates. Approach~A: constant model-input length, increasing real-time context. Approach~B: constant real-time context, decreasing temporal resolution.}
\vspace{-4mm}
\label{tab:samplerate}
\small
\setlength{\tabcolsep}{3.5pt}
\begin{tabular}{llcrrrrr}
\toprule
\textbf{DS} & \textbf{App.} & \textbf{Hz} & \textbf{Samples} &
  \textbf{Context} &
  \makecell{\textbf{Acc.}\\\textbf{(\%)}} &
  \makecell{\textbf{F1}\\\textbf{(\%)}} \\
\midrule
MH & A & 50 (base)  & 200 & 4\,s  & 96.68 & 96.22 \\
MH & A & 25         & 200 & 8\,s  & 96.36 & 95.50 \\
MH & A & 12.5       & 200 & 16\,s & 95.73 & 94.85 \\
MH & A & 6.25       & 200 & 32\,s & 94.58 & 85.31 \\
\cmidrule(lr){1-7}
MH & B & 50 (base)  & 200 & 4\,s  & 96.68 & 96.22 \\
MH & B & 25         & 100 & 4\,s  & 96.37 & 95.90 \\
MH & B & 12.5       &  50 & 4\,s  & 96.04 & 95.61 \\
MH & B & 6.25       &  25 & 4\,s  & 93.99 & 93.00 \\
\midrule
USC & A & 100 (base) & 500 & 5\,s  & 86.92 & 81.01 \\
USC & A & 50         & 500 & 10\,s & 90.29 & 83.87 \\
USC & A & 25         & 500 & 20\,s & 92.96 & 78.47 \\
USC & A & 12.5       & 500 & 40\,s & 90.12 & 50.83 \\
\cmidrule(lr){1-7}
USC & B & 100 (base) & 500 & 5\,s  & 86.92 & 81.01 \\
USC & B & 50         & 250 & 5\,s  & 86.66 & 80.74 \\
USC & B & 25         & 125 & 5\,s  & 85.52 & 78.52 \\
USC & B & 12.5       &  62 & 5\,s  & 85.31 & 78.26 \\
\bottomrule
\end{tabular}
\vspace{-4mm}
\end{table}

The results reveal qualitatively different behaviour across datasets.

For MHEALTH, Approach~B is consistently superior, losing fewer than 0.35 accuracy points at 25\,Hz ($\times$2 downsampling) and maintaining above 96\% at 12.5\,Hz ($\times$4). Approach~A's macro F1 collapses at 6.25\,Hz (85.31\%) because 32-second context windows span multiple activity bouts, corrupting majority-vote labels. We conclude that \solution{} operates reliably on MHEALTH-class sensors at \textit{25\,Hz or above}.

For USC-HAD, the picture is strikingly different. Approach~A at 50\,Hz ($\times$2 downsampling) produces a \emph{3.37-point accuracy
improvement} over the 100\,Hz baseline (90.29\% vs.\ 86.92\%), because the fixed 500-sample window now covers 10\,seconds rather than 5, providing
additional temporal context that more than compensates for halved frequency resolution. Beyond 25\,Hz, F1 collapses (50.83\% at 12.5\,Hz) as activity bouts are subsumed within windows. Approach~B degrades gracefully, losing only 1.61 accuracy points at 25\,Hz and stabilising around 85.3\% below that. This has a direct hardware implication: a single-IMU deployment targeting USC-HAD-class activities can \textit{reduce its sampling rate from 100\,Hz to 50\,Hz without accuracy loss and with measurable improvement}, simply by using the longer temporal context. For MHEALTH-class multi-sensor platforms, 25\,Hz is the practical lower bound for reliable operation.

\section{Extended Ablation Study}
\label{app:ext-ablation}

This section reports the remainder of the ablation study in Section~\ref{sec:ablation}. Table~\ref{tab:ext-ablation} extends the ablation in Table~\ref{tab:ablation}.

\begin{table*}[t]
\centering
\caption{Ablation study across all design axes (5-fold LOSO). Bold rows denote the default configuration for each group. }
\vspace{-4mm}
\label{tab:ext-ablation}
\small
\setlength{\tabcolsep}{4pt}
\begin{tabular}{cll cc cc c}
\toprule
\textbf{Grp} & \textbf{ID} & \textbf{Variant} &
  \multicolumn{2}{c}{\textbf{MHEALTH}} &
  \multicolumn{2}{c}{\textbf{USC-HAD}} &
  \textbf{Trainable} \\
\cmidrule(lr){4-5}\cmidrule(lr){6-7}
 & & &
  \makecell{\textbf{Acc.}\\\textbf{(\%)}} &
  \makecell{\textbf{F1}\\\textbf{(\%)}} &
  \makecell{\textbf{Acc.}\\\textbf{(\%)}} &
  \makecell{\textbf{F1}\\\textbf{(\%)}} &
  \textbf{Params} \\
\midrule
\multirow{3}{*}{D}
  & \textbf{D1} & \textbf{Frozen LLM (default)}      &
    \textbf{96.68} & \textbf{96.22} & \textbf{86.92} & \textbf{81.01} & \textbf{14.34\,M} \\
  & D2 & Full LLM fine-tuning                        & 96.36 & 95.80 & 85.40 & 78.02 & 138.78\,M \\
  & D3 & Partial fine-tune (last 2 layers)            & 95.95 & 95.25 & 87.08 & 80.82 & 28.51\,M \\
\midrule
\multirow{2}{*}{E}
  & \textbf{E1} & \textbf{Activity-name labels (default)} & \textbf{96.68} & \textbf{96.22} & 86.92 & \textbf{81.01} & \textbf{14.34\,M} \\
  & E2 & Generic class-ID labels                     & 96.34 & 95.72 & \textbf{87.19} & 80.79  & 14.34\,M \\
\midrule
\multirow{4}{*}{F}
  & F1 & \textbf{GPT2 (default)}         & \textbf{96.68} & \textbf{96.22} & \textbf{86.92} & \textbf{81.01} & \textbf{14.34\,M (frozen -117\,M)} \\
  & F2 & distill-gpt2                    & 96.12 & 95.83 & 85.71 & 79.43 & 14.34\,M (frozen - 82\,M) \\
  & F3 & gpt2-medium                     & 97.12 & 96.48 & 87.81 & 82.53 & 14.34\,M (frozen - 345\,M) \\
  & F4 & gpt2-large                     & 97.97 & 97.28 & 88.10 & 83.81 & 14.34\,M (frozen - 774\,M) \\
  & F5 & facebook-opt-125m               & 96.72 & 95.93 & 85.98 & 80.16 & 14.34\,M (frozen - 125\,M) \\
  & F6 & Qwen2.5-0.5B              & 96.43 & 95.45 & 82.37 & 79.09 & 14.34\,M (frozen - 494\,M) \\
  & F7 & TinyLlama-1.1B       & 96.10 & 95.16 & 81.74 & 78.25 & 14.34\,M (frozen - 1.1\,B) \\
\midrule
\multirow{2}{*}{G}
  & \textbf{G1} & \textbf{GPT2 (default)} & \textbf{96.68} & \textbf{96.22} & \textbf{86.92} & \textbf{81.01} & \textbf{14.34\,M} \\
  & G2 & linear-head                     & 93.14 & 91.57 & 82.11 & 75.84 & 14.63\,M \\
\midrule
\multirow{2}{*}{H}
  & \textbf{H1} & \textbf{with task prompt} & \textbf{96.68} & \textbf{96.22} & \textbf{86.92} & \textbf{81.01} & \textbf{14.34\,M} \\
  & H2 & without prompt, sensor only          & 94.89 & 93.67 & 82.79 & 76.71 & 14.34\,M \\
\bottomrule
\end{tabular}
\vspace{-4mm}
\end{table*}

\subsection{\textbf{LLM Training Strategy (Group~D)}}
\label{sec:abl-llm}

The frozen LLM strategy (D1) consistently outperforms or matches all fine-tuning variants across both datasets, and this result is \emph{strengthened} on the harder USC-HAD benchmark. On MHEALTH, full fine-tuning (D2) costs 0.32\,accuracy points relative to frozen. On USC-HAD, the penalty for full fine-tuning grows to \textit{2.99\,F1 points}, despite expanding trainable parameters 9.7-fold (138.78\,M vs.\ 14.34\,M) and increasing per-epoch training time from 13.3\,s to 18.1\,s. Partial fine-tuning of the last two transformer blocks (D3) offers a mixed result: slightly above D1 on USC-HAD accuracy (87.08\% vs.\ 86.92\%) but below D1 on F1 (80.82\% vs.\ 81.01\%), and substantially below D1 on MHEALTH. The inconsistency of D3 across datasets, combined with its higher parameter count and less pronounced per-epoch speedup (9.9\,s/epoch vs.\ 13.3\,s for D1 on USC-HAD), makes it an unreliable choice.

The mechanism behind frozen LLM superiority is consistent across both datasets: the pretrained GPT-2 assigns higher prior probability to
semantically coherent activity-name completions than to random token sequences, and this prior regularises the classification under NLL-based
scoring. Fine-tuning erodes this prior. The penalty is larger on USC-HAD because activity signals are more ambiguous with only 6 channels, making
the LLM's pretrained prior play a relatively larger role in resolving borderline predictions. \solution{} therefore maintains the LLM as a frozen
reasoning engine on all platforms and datasets.

\subsection{\textbf{Semantic Label Representation (Group~E)}}
\label{sec:abl-labels}

The effect of semantic label representation is dataset-dependent and proportional to sensor richness. On MHEALTH, activity-name labels (E1)
outperform generic class identifiers (E2) by 0.34\,accuracy points and 0.50\,F1 points, confirming that the LLM's world knowledge about human
activities provides a meaningful discriminative signal when 21-channel sensor data offers sufficient discriminability. On USC-HAD, the gap
narrows substantially: E2 is 0.27\,accuracy points above E1 but 0.22\,F1 points \emph{below} E1, with neither variant consistently superior across
folds. This suggests that when sensor modalities are sparse (6 channels, single IMU) and inter-class NLL margins are narrower and less reliable, the LLM's semantic priors offer diminishing benefit: activities such as walking left and walking right are not easily resolved by their names even
with GPT-2's full world model. The results establish that semantic label benefit scales with sensor richness, and that \solution{} remains
fully functional with generic class identifiers in low-modality settings.

\subsection{\textbf{Robustness to LLM Backbone Choice (Group F).}}
Group F evaluates the robustness of \solution{} to different LLM backbones while keeping the sensor tokenizer and the 16-token representation fixed. The results show that performance is stable across a range of causal language models, with consistent gains observed when scaling within the GPT family. 
However, replacing GPT with other architectures (OPT, Qwen, TinyLlama) does not yield consistent gains despite comparable or larger parameter counts. This indicates that performance depends not only on model size but also on alignment between the tokenizer output and the LLM architecture.
Overall, these results demonstrate that \solution{} is robust to the choice of LLM backbone, while also benefiting from increased model capacity within a well-aligned family. GPT-2 provides the best efficiency--performance trade-off, while larger GPT variants offer additional accuracy when computational budget permits.

\subsection{\textbf{LLM  vs.\ Linear Classification Head (Group~G)}}
\label{sec:classification_head}

Group~G tests whether the frozen GPT-2 scoring mechanism can be replaced by a standard linear classifier. Although the linear-head variant uses a comparable number of trainable parameters to the default configuration, it performs substantially worse on both datasets. Compared with the linear head, GPT-2 scoring improves F1 by 4.65 points on MHEALTH and 5.17 points on USC-HAD. This confirms that the benefit of \solution{} does not come only from the sensor tokenizer: the frozen LLM provides an additional discriminative prior that improves robustness across both rich multi-sensor and sparse single-IMU settings. Therefore, \solution{} retains the frozen GPT-2 scoring head as the default prediction mechanism.

\subsection{\textbf{Task Prompt Conditioning (Group~H)}}
\label{sec:task_prompt}

Group~H evaluates whether the task prompt provides useful conditioning beyond the latent sensor tokens alone. Compared with the sensor-only variant, adding the task prompt improves F1 by 2.55 points on MHEALTH and 4.30 points on USC-HAD, while increasing inference latency by only 1.83\,ms. This shows that the prompt helps the frozen LLM interpret the sensor-token sequence in the intended HAR classification context, with a negligible latency cost. We therefore retain task-prompt conditioning as part of the default input formulation.